\documentclass[]{bytedance_seed}

\usepackage[toc,page,header]{appendix}

\usepackage{amsmath} 
\usepackage{amssymb}   
\usepackage{mathtools}  
\usepackage{amsfonts}   
\usepackage{algorithm}
\usepackage{algorithmic}

\title{Hydra-Nav: Object Navigation via Adaptive Dual-Process Reasoning}

\author[1,2,*]{Zixuan Wang}
\author[1]{Huang Fang}
\author[1,3,*]{Shaoan Wang}
\author[1]{Yuanfei Luo}
\author[1]{Heng Dong} 
\author[1, \dagger]{\protect\\ Wei Li}
\author[4, \dagger]{Yiming Gan}

\affiliation[1]{ByteDance Seed}
\affiliation[2]{Institute of Automation, Chinese Academy of Sciences}
\affiliation[3]{Peking University}
\affiliation[4]{Institute of Computing Technology, Chinese Academy of Sciences}

\contribution[*]{Work done at ByteDance Seed}
\contribution[\dagger]{Corresponding authors}

\abstract{
While large vision-language models (VLMs) show promise for object goal navigation, current methods still struggle with low success rates and inefficient localization of unseen objects—failures primarily attributed to weak temporal-spatial reasoning.
Meanwhile, recent attempts to inject reasoning into VLM-based agents improve success rates but incur substantial computational overhead. 
To address both the ineffectiveness and inefficiency of existing approaches, we introduce \proj{}, a unified VLM architecture that adaptively switches between a deliberative ``slow system'' for analyzing exploration history and formulating high-level plans, and a reactive ``fast system'' for efficient execution. 
We train \proj{} through a three-stage curriculum: (i) spatial-action alignment to strengthen trajectory planning, (ii) memory-reasoning integration to enhance temporal-spatial reasoning over long-horizon exploration, and (iii) iterative rejection fine-tuning to enable selective reasoning at critical decision points. 
Extensive experiments demonstrate that \proj{} achieves state-of-the-art performance on the HM3D, MP3D, and OVON benchmarks, outperforming the second-best methods by 11.1\%, 17.4\%, and 21.2\%, respectively. 
Furthermore, we introduce SOT (\textbf{S}uccess weighted by \textbf{O}peration \textbf{T}ime), a new metric to measure search efficiency across VLMs with varying reasoning intensity. 
Results show that adaptive reasoning significantly enhances search efficiency over fixed-frequency baselines. 
}

\date{\today}
\correspondence{\email{liwei.85@bytedance.com}, \email{ganyiming@ict.ac.cn}}

\checkdata[Project Page]{\url{https://zixuan-wang99.github.io/Hydra-Nav/}}

\ifx\figurename\undefined \def\figurename{Figure}\fi
\renewcommand{\figurename}{Fig.}
\renewcommand{\paragraph}[1]{\textbf{#1} }

\newcommand\hl{\bgroup\markoverwith
  {\textcolor{yellow}{\rule[-.5ex]{2pt}{2.5ex}}}\ULon}

\newcommand{\proj}{\textsc{Hydra-Nav}}
\newcommand{\projsft}{\textsc{Hydra-Nav-SFT}}
\newcommand{\projrft}{\textsc{Hydra-Nav-IRFT}}

\newcommand{\no}[1]{#1}
\renewcommand{\no}[1]{}
\newcommand{\RNum}[1]{\uppercase\expandafter{\romannumeral #1\relax}}

\begin{document}
\maketitle


\section{Introduction}
\label{Introduction}

Object goal navigation is an emerging embodied task that requires a robot to actively explore the physical world and locate target objects using only egocentric perception \cite{chaplot2020object, sun2024survey}. Object navigation has attracted considerable attention from both academia and industry, as the ability to search for objects is essential for deploying robots in home environments. This task demands a combination of multiple core capabilities: (i) strong spatial reasoning to understand the spatial structure of the environment; (ii) effective long-term memory to retain information about explored scenes; (iii) robust long-horizon trajectory planning to explore the environment \cite{habibpour2025history,liu2025aligning}. 

The rise of large vision-language models (VLMs) has significantly advanced object navigation. Recent works \cite{yu2023l3mvn,nasiriany2024pivot,yokoyama2024vlfm,goetting2024end,zhong2025p3nav,zhang2025embodied} leverage VLMs to improve navigation success rates, benefiting from their strong generalization capabilities. Despite these advances, current methods still fall short of human-level performance. This gap primarily stems from the limited temporal-spatial understanding ability of the underlying VLMs. For instance, the model often fails to maintain a coherent memory of visited regions, leading to redundant exploration of the same areas. To narrow the gap, recent works incorporate memory mechanisms \cite{zhang2024navid,yang20253d,zhang2024uni} and explicit reasoning capabilities \cite{nie2025wmnav,zhu2025move,cao2025cognav,li2025compassnav,cai2025cl,qiao2025open,liu2025nav,xiang2025nav} at each inference step. However, their reasoning process tends to 
focus on current observations, overlooking the analysis of historical trajectory and progress of exploration. Furthermore, while frequent reasoning improves success rates, it leads to unnecessary inference costs during routine tasks. For instance, walking down a long, unobstructed hallway yields highly consistent visual inputs, which are best handled by simple reactive control.

We introduce \proj{}, a VLM tailored for object navigation. To improve the temporal-spatial reasoning ability of the base VLM, we develop a data synthesis pipeline and collect 500K trajectories with high-quality reasoning from simulated environments. We carefully design the reasoning synthesis prompt to include three components: reviewing historical memory, analyzing current observation, and planning future trajectory based on both. The base VLM (Qwen2.5-VL-7B) trained on our data demonstrates strong temporal-spatial reasoning ability and achieves state-of-the-art performance on widely-used object navigation benchmarks.
To reduce the overhead of frequent reasoning, we design \proj{} as a slow-fast system that allows triggering reasoning at arbitrary frequency. Meanwhile, we use iterative rejection-sampling finetuning (IRFT) to teach \proj{} to adaptively switch between the slow and fast modes. IRFT not only increases search efficiency but also boosts the success rate across all settings. Our contributions are summarized as follows:

\begin{itemize}
    \item We propose \proj{}, a navigation agent that supports adaptive temporal-spatial reasoning within a single VLM architecture. Unlike prior approaches that trigger reasoning at a fixed frequency, \proj{} learns to adaptively switch between a strategic slow system and a reactive fast system, reducing reasoning overhead while maintaining a high success rate.

    \item We design a curriculum training pipeline to progressively enhance \proj{}'s spatial-action alignment (\Cref{sec:stage1}), temporal-spatial reasoning (\Cref{sec:stage2}), and adaptive reasoning ability (\Cref{sec:stage3}).
    
    \item We conduct extensive experiments on widely-used object navigation benchmarks (HM3D, MP3D, and OVON). \projsft{} with fixed reasoning frequency achieves a $50.9\%$ success rate on the OVON Val-Unseen benchmark, $5.7\%$ higher than the previous state-of-the-art. \projrft{} achieves state-of-the-art success rate on HM3D, MP3D, and OVON, outperforming the second-best method by 11.1\%, 17.4\%, and 22.3\% respectively. Furthermore, we introduce a new metric, SOT (\textbf{S}uccess weighted by \textbf{O}peration \textbf{T}ime), to evaluate the search efficiency of navigation agents with different reasoning intensity. We show that \projrft{} achieves better SOT than \projsft{}, demonstrating the effectiveness of IRFT.

\end{itemize}
\section{Related Work}

\begin{figure*}[ht]  
    \centering      
    \includegraphics[width=\textwidth]{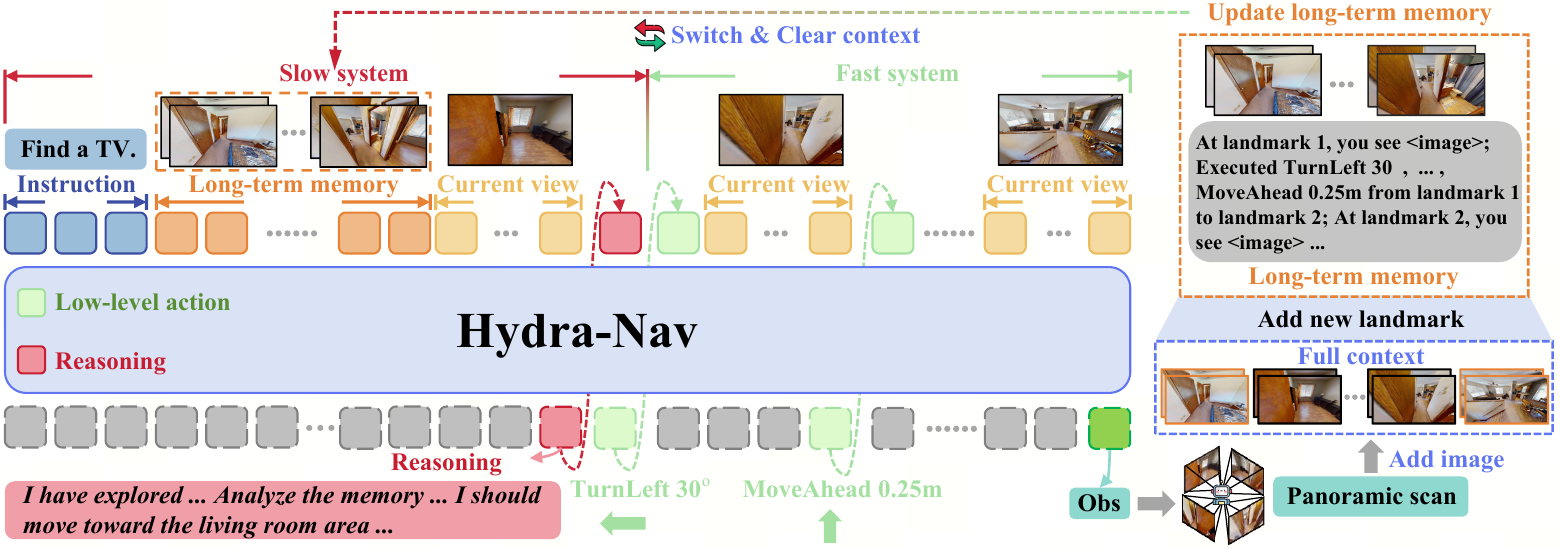}
    \caption{\textbf{The architecture of \proj{}.} \proj{} receives user instruction, long-term memory, and previous image-action pairs, then outputs reasoning (optionally) and meta-actions. \proj{} adaptively switches between the fast and slow systems by outputting the special transition token \texttt{obs}.  Specifically, the panoramic scan triggered by \texttt{obs} is extracted as a new landmark and inserted into the existing long-term memory.}
    \label{navi-agent framework}
\end{figure*}

\subsection{Vision Language Models for Object Navigation}

Incorporating vision language models (VLMs) into object navigation has shifted the field from traditional methods—ranging from end-to-end reinforcement learning policies \cite{zhu2017target,yang2018visual,lyu2022improving,druon2020visual,wijmans2019dd} to modular map-based planning \cite{chaplot2020object,chaplot2020learning,sun2026dadu,ramakrishnan2022poni,chaplot2020neural}—toward leveraging pre-trained knowledge. VLMs-based methods enable agents to interpret high-level instructions and perform common sense reasoning, facilitating robust open-world understanding and generalization. 

VLM-based navigation agent generally follows three paradigms. First, modular agent system approaches \cite{yu2023l3mvn,nasiriany2024pivot,yokoyama2024vlfm,goetting2024end,zhou2025beliefmapnav,yang20253d,Wang2026VLingNavEN} adopt a hierarchical framework where the VLM acts as a global policy to select waypoints, leaving execution to a deterministic local planner. Recent works \cite{nie2025wmnav,zhang2025nava,long2024instructnav,zhu2025move,cao2025cognav,li2025compassnav} further enhance this by using Chain-of-Thought (CoT) for deeper scene analysis and planning. Second, end-to-end methods \cite{zhang2024uni,gao2025octonav,zhang2024navid,zhong2025p3nav,zhang2025embodied,liu2025nav,xiang2025nav} fine-tune VLMs to directly predict low-level control actions from visual inputs. Third, dual-system architectures \cite{wei2025ground,xue2025omninav} employ a ``slow-fast'' paradigm, where a VLM generates periodic sub-goals to guide a lightweight, high-frequency controller.

However, a critical trade-off between reasoning depth, inference efficiency, and system unity persists across these paradigms. Modular agent system and end-to-end approaches incur prohibitive latency by applying CoT reasoning at every control step. While dual system architectures mitigate this via hierarchical ``slow-fast'' paradigm, they suffer from architectural fragmentation and rigid, heuristic switching, lacking the flexibility to adaptively reasoning. In contrast, \proj{} unifies the dual-process mechanism within a single VLM, enabling adaptive switching between direct low-level execution and CoT reasoning.

\subsection{Adaptive Reasoning}
Adaptive thinking dynamically selects between direct response and step-by-step CoT thinking \cite{chen2024toward,chung2025thinker,wang2025adareasoner,wu2025efficiency,wu2025arm,yu2025think,zhan2025kat,zhang2025adaptthink}. Existing methods \cite{peng2025counterfactual,luo2025adathinkdrive} typically rely on rule-based heuristics to categorize scene complexity, employing supervised fine-tuning or reinforcement learning to learn when to think. Besides, Game-tars \cite{wang2025game} utilizes rejection fine-tuning on extensive human-annotated datasets to mimic human-like adaptive reasoning. However, in the domain of object navigation, manually demarcating simple versus challenging scenarios is non-trivial, and large-scale human demonstrations of adaptive reasoning are scarce. Crucially, unlike recent dual-process approaches that rely on modular pipelines \cite{zhong2025run,yue2025think,zhou2025fsr}, \proj{} pioneers the unification of adaptive slow-fast reasoning within a single VLM architecture for object navigation. To address this, we introduce iterative rejection fine-tuning (IRFT). Instead of relying on external supervision, we leverage on-policy rollouts to autonomously finding ``stagnation points''—moments where reactive low-level action policies fail. By injecting reasoning retrospectively at these points, we train the model to trigger CoT adaptively when necessary.

\section{Method}

\subsection{Problem Formulation}

We formulate object goal navigation as a partially observable markov decision process (POMDP). At each time step $t$, the agent receives an egocentric observation $o_t$ and a natural language instruction $I_{goal}$ specifying the target object. Based on the history ${H}_t$, the agent generates an output $y_{t}$ from a hybrid action space $\mathcal{A} = \mathcal{A}_{motor} \cup \mathcal{A}_{reason}$. The motor action space $\mathcal{A}_{motor}$ comprises discrete locomotion primitives (e.g., $\texttt{MoveAhead 0.25m}$) and a specialized system transition token $obs$ to decide if switch to the slow system. The reasoning space $\mathcal{A}_{reason}$ consists of natural language tokens used by the slow system to generate the reasoning text. This unified formulation allows a VLM to flexibly alternate between executing low-level control and generating high-level strategic guidance.

\subsection{The Dual-process System}
\label{subsec:overview}

\begin{figure}[t]  
    \centering      
    \includegraphics[width=0.8\textwidth,]{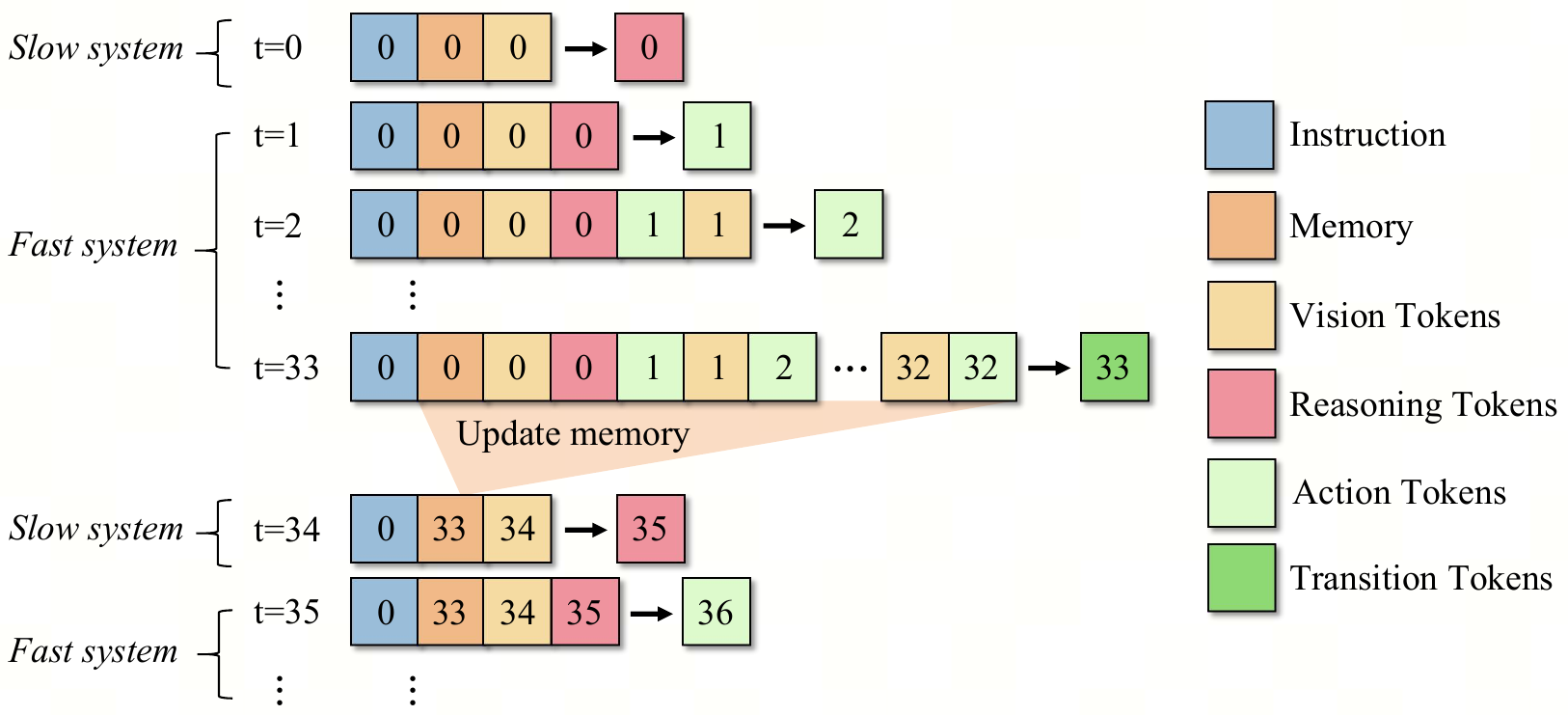}
    \caption{\textbf{Illustration of the context organization of \proj{} during inference.} The context starts with a system prompt containing the user instruction and long-term memory. Short-term memory is organized as interleaved image-action pairs. When a transition token is encountered, we update the memory and clear the image-action pairs. Note that each token block in the figure represents a sequence of multiple tokens.}
    \label{fig:inference_token}
\end{figure}

As illustrated in \Cref{navi-agent framework}, \proj{} unifies high-level planning and low-level meta actions within a single VLM architecture. Unlike prior hierarchical approaches that rely on separate models \cite{ding2025adanav,nie2025wmnav}, \proj{} operates as an end-to-end dual-process system. 
Below, we detail a full cycle of the dual process; see \Cref{fig:inference_token} for an illustration.

\paragraph{Slow system.} The cycle initiates with the slow system, responsible for global localization and high-level temporal-spatial reasoning. The input consists of the goal instruction (e.g., ``Find a TV.''), the current panoramic observation comprising 4 RGB images obtained by rotating the head camera at $90^\circ$ intervals ($0^\circ, 90^\circ, 180^\circ, 270^\circ$), and a structured long-term memory. We construct the memory as a serialized graph of text-image interleaved landmark nodes connected by action edges (see \Cref{Memory Example}). The model generates a reasoning text to analyze exploration progress followed by an immediate subsequent meta action (e.g., $\texttt{<think\_start>}$ \ldots move toward the living room area  \ldots $\texttt{<think\_end>}$\texttt{MoveAhead 0.25m}).

\paragraph{Fast system.} Upon generating the high-level plan, the model transitions to the fast system. This phase operates as a multi-turn dialogue with long-term memory and reasoning text from the previous slow system as the system prompt. For subsequent steps, the model takes the dialogue history and the current observation frame as input. Utilizing KV-caching, the model encodes only the latest egocentric frame and autoregressively decodes low-level atomic actions (e.g., $\texttt{MoveAhead 0.25m}$). This design avoids reprocessing the full history context, thereby saving computation.

\paragraph{Adaptive switching mechanism.} The transition from the fast system back to the slow system is self-triggered. When the agent completes a sub-goal in the high-level plan, or when the current observation invalidates the existing plan, the model outputs a special token \texttt{obs}. This token initiates a panoramic scan (four $90^\circ$ rotations) to capture new observations and construct a new landmark node. The node is then connected to the previous landmark through the executed action sequence and added to the long-term memory. To keep the context length manageable, the memory is dynamically pruned: the start and end nodes are always preserved, while intermediate landmarks are uniformly sampled, with a maximum of 10 landmarks retained. With the updated memory, the agent re-enters the slow system to revise its high-level plan.

Next, we describe our curriculum training pipeline that progressively enhance the spatial-action alignment, temporal-spatial reasoning and adaptive reasoning of \proj{}.

\subsection{Stage 1: Spatial-action Alignment}
\label{sec:stage1}
The goal of this stage is to collect a large number of trajectories to train the base VLM into a navigation agent. Following recent works \cite{zhang2024uni,liu2025nav}, we use annotated data from the training sets of HM3D \cite{{yadav2023habitat}}, MP3D \cite{chang2017matterport3d}, and OVON \cite{yokoyama2024hm3d} to generate 500K trajectories using the $A^*$ planner. Each trajectory is formatted as a multi-turn conversation:
$$\tau \coloneqq \left\{ I^{(1)}_{\text{sys}}, ({v}_1, {a}_1), ({v}_2, {a}_2), \ldots, ({v}_T, {a}_T) \right\},$$where $I^{(1)}_{\text{sys}}$ represents the system instruction used in this stage, $v_t$ denotes the egocentric RGB observation at time step $t$, $a_t \in \{ \texttt{MoveAhead 0.25m}, \texttt{TurnLeft 30}^{\circ}, \allowbreak \texttt{TurnRight 30}^{\circ}, \texttt{End} \}$ is the meta action executed at time $t$. 
To reduce computational cost, we compute the loss over all turns in a single forward-backward pass, allowing each trajectory to be processed in one operation. The training parameters used by this stage is shown in \Cref{tab:training_params}.

\subsection{Stage 2: Reasoning-memory Integration}
\label{sec:stage2}

\begin{figure*}[t]  
    \centering      
    \includegraphics[width=\textwidth]{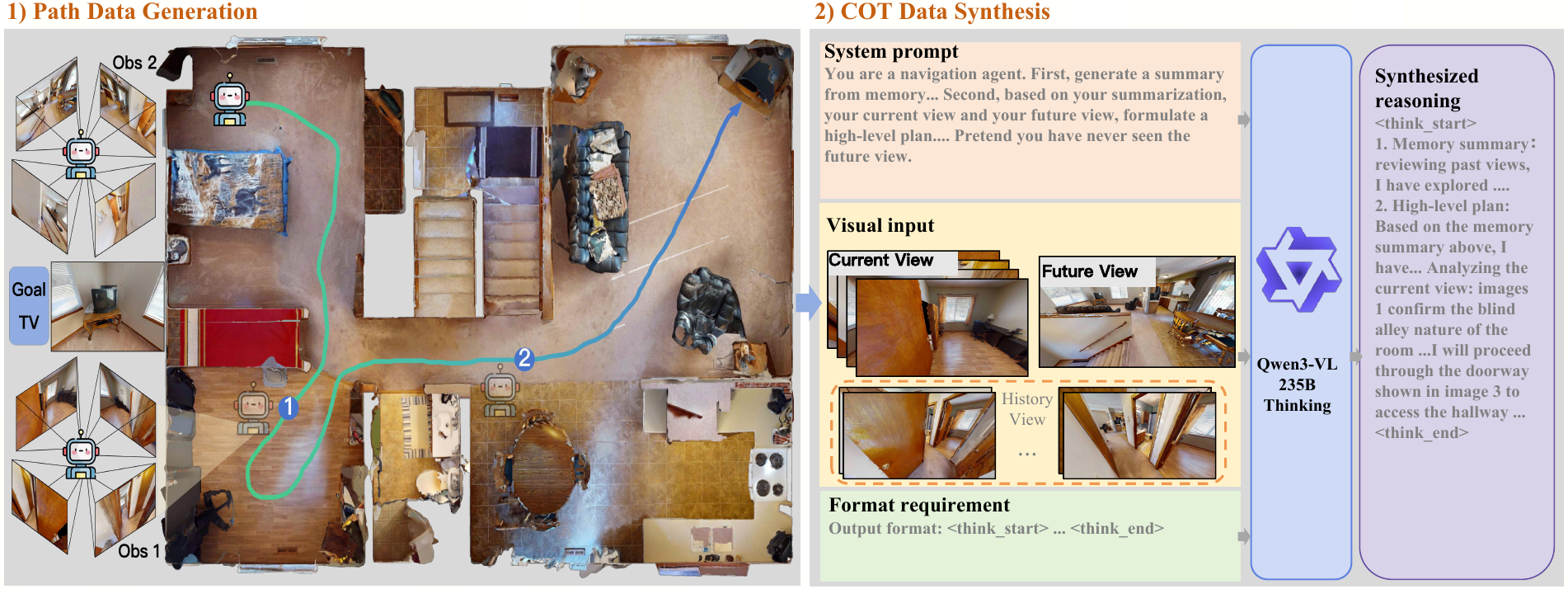}
    \caption{\textbf{An illustration of the data synthesis pipeline used in stage 2.} The left side shows our trajectory generation strategy, where the robot visits Point 1 and Point 2 before reaching the goal. The right side illustrates how we prompt Qwen3-VL-235B-Thinking to produce high-quality reasoning traces. }
    \label{fig:cot_data_generation_v2}
\end{figure*}

After Stage 1, the base VLM is able to generate reasonable, collision-free trajectories in the simulated environment, but lacks the ability to explore the scene when its initial attempt fails. Stage 2 equips the model with the ability to (i) explore the environment while recovering from failures, and (ii) reason on the long-term memory and plan for future exploration accordingly. Next, we describe how we enable the model to acquire these two capabilities.

\paragraph{Trajectory generation.} Unlike Stage 1, which uses $A^*$ to generate trajectories, here we construct trajectories using a heuristic waypoint selection algorithm. Specifically, for each scene, we define a score function ${S}(p; p_{\text{init}})$ over all reachable points ${p} \in \mathbb{R}^3$, conditioned on the robot's initial position $p_{\text{init}}$ (see \Cref{sec:appendx_score_function} for details). For each initial position, we select the two highest-scoring points: $p_1, p_2 = \text{top-2}_{p \in \mathbb{R}^3} S(p; p_{\text{init}})$. The trajectory is then constructed as $p_{\text{init}} \to p_1 \to p_2 \to p_{\text{target}}$ or $p_{\text{init}} \to p_2 \to p_1 \to p_{\text{target}}$, depending on the total length of the path, where $p_{\text{init}}$ denotes the starting position and $p_{\text{target}}$ represents the location of the target object. As shown in \Cref{fig:cot_data_generation_v2}, the robot first visits these two high-scoring exploration points (Point 1 and Point 2) before finally reaching the target (e.g., a television).

\paragraph{Temporal-spatial reasoning synthesis.} After generating trajectories with exploration described above, we partition each trajectory into a set of segments with fixed length (we set the length to be 16). Then we
insert memory and reasoning text at the beginning of each segment, we also insert the transition token \texttt{obs} at the end of each segment. Each segment is formatted as a multi-turn conversation: 
\begin{align*}
    s_1 &\coloneqq \{ I^{(2)}_{\text{sys}}, (m_1, v_1, r_1, a_1), (v_2, a_2), \ldots, (v_{16}, \texttt{obs}) \} \\
    s_2 &\coloneqq \{ I^{(2)}_{\text{sys}}, (m_2, v_{17}, r_2, a_{17}), (v_{18}, a_{18}), \ldots, (v_{32}, \texttt{obs}) \} \\
    & \ldots
\end{align*}
where $s_i$ denote the $i$-th segment of the trajectory, $m_i$ is the long-term memory (a compression of previous segments, see \Cref{Memory Example} for the details), $r_i$ is the reasoning text of the $i$-th segment.

We synthesize reasoning text with the following procedure: at time
$t$ that triggers the slow system, we first prompt Qwen3-VL-235B Thinking \cite{Qwen3-VL} to review historical images and summarize the search progress so far. Then, we feed the summary, together with the current view and the future view (the panoramic scan at the beginning of the next segment) to the model. We ask it to produce reasoning text that serves to guide the future search plan for the target object. To prevent information leakage, we explicitly prompt the model to treat the future view only as a hidden guide, ensuring it is not mentioned in the output. Furthermore, we use a LLM to verify and filter out any reasoning text that inadvertently reveal future information.
The prompt used to synthesize reasoning is shown in \Cref{Reasoning Prompt}. See the right side of \Cref{fig:cot_data_generation_v2} for an illustration of the reasoning synthesis workflow. 

After reasoning synthesis, we can no longer process a full trajectory within one forward-backward pass, because different segments from the same trajectory do not share the same context. As a result, we train the model at the segment level, and the training efficiency of stage 2 is lower than that of stage 1. In addition to the navigation data, we also mix some visual question answering data into stage 2 training to maintain the general capability of the base VLM.

\subsection{Stage 3: Adaptive Reasoning via Iterative Rejection Fine-tuning}
\label{sec:stage3}

While stage 2 establishes the fundamental mechanics of the dual-system architecture, it switches between the slow and fast systems at a fixed frequency. This static switching policy can incur unnecessary computational overhead in trivial scenarios and often fails to trigger the slow system at critical ``stagnation points'' where navigation collapses. 

We first define ``stagnation points'' as moments where the agent fails to produce effective exploration progress. We identify two specific types of ``stagnation point'':

\begin{enumerate}
    \item Repetitive exploration: The agent exhibits repetitive exploration at time $t$ if it revisits a position from at least $T_{\text{stag}}$ steps earlier, i.e., $\exists k \in [0, t-T_{\text{stag}}]$ such that $\| p_t - p_k \| \leq \delta_{\text{stag}}$.
    \item Lack of progress: The agent lacks progress at time $t$ if $\text{dist}(p_t, G) > \text{dist}(p_{t - \Delta t}, G)$, where $\Delta t \sim \mathcal{U}(20, 35)$. 
    We choose the distribution $\mathcal{U}(20, 35)$ for two reasons: (i) uniform sampling increases data diversity and mitigates overfitting risk; (ii) the range $[20,35]$ is large enough to avoid false positives when the agent navigates around obstacles.
\end{enumerate}
We set $T_{\text{stag}}=20$ and $\delta_{\text{stag}}=0.5$m. These conditions determine when the reactive fast system requires intervention from the reasoning (slow) system.

To collect training data, we conduct on-policy rollouts where the agent operates primarily in the fast system and triggers the slow system at the defined ``stagnation points'', see \Cref{alg:rollout_stage3} for details. We retain trajectories where the agent successfully reaches the goal. 
Otherwise, we repair failed trajectories using different strategies based on their failure modes.

We observe that failed trajectories fall into one of two categories:
(i) \textbf{Timeout}: The episode length exceeds the maximum step limit ($T > 200$); or (ii) \textbf{Target misidentification}: The agent misidentifies the target at time $t$ if $a_t=\texttt{End}$ and $\text{dist}(p_t, G) > \delta_{\text{success}}$, where $\text{dist}(p_t, G)$ denotes the minimum distance between $p_t$ and the target object set $G$.

For each failure mode, we first determine the intervention timestamp $t^*$: for target misidentification, we set  $t^*$ to the final step; for timeout, we scan the trajectory for stagnation points and select the point where the agent was spatially closest to the target.

Once the intervention timestamp $t^*$ is identified, we use the following procedure to repair the trajectory:
\begin{itemize}
  \item We substitute the erroneous action $a_{t^*}$ with the transition token \texttt{obs}, thereby switching the model to the slow system at $t^*$.
  \item We construct a new trajectory by concatenating the history $p_{1:t^*}$ with the optimal path (generated via an $A^*$ planner) from $p_{t^*}$ to the target $p_{\text{target}}$.
  \item We keep the reasoning of the historical path $p_{1:t^*}$ and synthesize a new reasoning text for the corrective path $p_{t^*} \to p_{\text{target}}$ following the similar procedure described in stage 2.
  \item If the length of the new trajectory exceeds 400, we simply drop this trajectory.
\end{itemize}

The detailed data collection procedure is described in \Cref{alg:reapiar}. The trajectories produced by this procedure do not trigger reasoning at a fixed frequency, allowing the model trained on them to adaptively switch to the slow system. We iteratively apply this ``reject-and-repair'' data synthesis strategy using the latest checkpoint, progressively refining the model's adaptive reasoning capability. This pipeline is analogous to the rejection fine-tuning methods used in LLM training \cite{touvron2023llama,guo2025deepseek}, and we refer to the model at this stage as \proj{}-IRFT.

\subsection{Training Configurations}
\label{sec:appendx_training_config}

We use the AdamW optimizer ($\beta_1=0.9, \beta_2=0.95$) with bfloat16 precision and the cross-entropy loss for all three stages. A cosine learning rate scheduler is used with a warmup ratio of 0.1. We set the weight decay to 0.1 and gradient clipping to 1.0 for stability. The image inputs are fixed at $640 \times 480$ resolution with a max sequence length of 32k. Other training details are shown in \Cref{tab:training_params}.
\begin{table}[t]
\centering
    \begin{small}
    \caption{Hyperparameters for the three stage training pipeline.}
    \label{tab:training_params}
    \begin{tabular}{l|c|c|c}
        \toprule
        \textbf{Configuration} & \textbf{Stage 1} & \textbf{Stage 2} & \textbf{Stage 3} \\
        \midrule
        
        Dataset size & 500k trajectories & 565k mixed samples & $\sim$60k  
        trajectories (per round) \\
        Token count  & 20.1B & 8.3B & 4.5B \\
        Base model & {Qwen2.5-VL-7B} & Stage 1 model & Stage 2 model \\
        Hardware & 128 GPUs & 96 GPUs & 64 GPUs \\
        Global batch size & 1024 & 768 & 512 \\
        Epochs & 2 & 2 & 3 \\
        
        Training Cost & 140 hours & 100 hours & 50 hours \\

        \midrule

        Peak learning rate & $1 \times 10^{-5}$ & \multicolumn{2}{c}{$1 \times 10^{-6}$} \\
        Min learning rate & $1 \times 10^{-6}$ & \multicolumn{2}{c}{$1 \times 10^{-7}$} \\
        
        \midrule

    \end{tabular}
    \end{small}
\end{table}
\section{Experiments}

To evaluate the performance of \proj{}, we carry out extensive experiments on commonly used benchmarks. Specifically, we aim to address the following three research questions:
\begin{itemize}
    \item Does \proj{} outperform existing methods in terms of success rate and search efficiency on unseen objects and scenes? How does \proj{} behave at different training stages?
    \item How do different training strategies (e.g., with/without reasoning and memory modules) and data collection strategies affect the final performance of \proj{}?
    \item How does IRFT influence the final success rate and task completion time—which includes both robot execution time and model reasoning time?
\end{itemize}

\subsection{Experimental Setup}
\paragraph{Benchmarks.} We evaluate \proj{} on the HM3D \cite{yadav2023habitat}, MP3D \cite{chang2017matterport3d} for object goal navigation, and OVON \cite{yokoyama2024hm3d} for open-vocabulary object navigation.

\paragraph{Evaluation metrics.} We report two widely-used metrics: success rate (SR) and success weighted by path length (SPL). SR measures the proportion of successful episodes, while SPL further penalizes successful episodes by the ratio of the optimal path length to the actual path length. Our evaluation protocol is consistent with prior work \cite{zhang2024uni,zhu2025move} and follows standard practice.

While SPL is a standard metric for measuring path efficiency, it overlooks the computational cost incurred by reasoning models. To address this limitation, we introduce \textbf{S}uccess weighted by \textbf{O}peration \textbf{T}ime (SOT), a metric that penalizes successful episodes by the ratio of optimal search time to actual search time:
$$\text{SOT} \coloneqq \frac{1}{N} \sum_{i=1}^{N} S_i \frac{T_{\text{optimal}}}{T_{\text{actual}}}.$$
Here, $S_i \in \{0, 1\}$ is the success indicator, $T_{\text{optimal}}$ is the theoretical minimum physical time required to reach the target, while $T_{\text{actual}}$ represents the total operating time, comprising two components: (i) \textit{robot execution time}, computed based on execution time of meta actions (e.g., 1.0s for \texttt{MoveAhead});
(ii) \textit{model inference latency}, the time required for the VLM to generate reasoning and action tokens. See \Cref{detail for SOT} for more details.

\begin{table*}[t!]
\caption{\textbf{Comparison with state-of-the-art methods on HM3D, MP3D, and OVON benchmarks.} The attributes denote: \textbf{Low}: low-level action output, \textbf{High}: high-level planning/action output, \textbf{Dual}: dual-system architecture. \textbf{RGB}: uses RGB observations. \textbf{Depth}: uses Depth observations. The \textbf{best} and the \underline{second best} results are denoted by \textbf{bold} and \underline{underline}.}
\label{tab:sota_comparison}
\begin{center}
\begin{small}
\resizebox{\textwidth}{!}{
\begin{tabular}{l ccccc cc cc cc cc cc}
\toprule
\multirow{3}{*}{Method} & \multicolumn{2}{c}{Observation} & \multicolumn{2}{c}{Action} & \multirow{3}{*}{Dual} & \multicolumn{2}{c}{HM3D} & \multicolumn{2}{c}{MP3D} & \multicolumn{6}{c}{OVON} \\
\cmidrule(lr){2-3} \cmidrule(lr){4-5} \cmidrule(lr){7-8} \cmidrule(lr){9-10} \cmidrule(lr){11-16}
 & \multirow{2}{*}{RGB} & \multirow{2}{*}{Depth} & \multirow{2}{*}{Low} & \multirow{2}{*}{High} & & \multicolumn{2}{c}{Val} & \multicolumn{2}{c}{Val} & \multicolumn{2}{c}{Val-Seen} & \multicolumn{2}{c}{Val-Synonyms} & \multicolumn{2}{c}{Val-Unseen} \\
\cmidrule(lr){7-8} \cmidrule(lr){9-10} \cmidrule(lr){11-12} \cmidrule(lr){13-14} \cmidrule(lr){15-16} 
 & & & & & & SR & SPL & SR & SPL & SR & SPL & SR & SPL & SR & SPL \\
\midrule
VoroNav \cite{wu2024voronav}     & $\surd$ & $\surd$ & & $\surd$ & & 42.0 & 26.0 & - & - & - & - & - & - & - & - \\
InstructNav \cite{long2024instructnav}  & $\surd$ & $\surd$ & & $\surd$ & & 50.0 & 20.9 & - & - & - & - & - & - & - & - \\
VLMnav \cite{goetting2024end}    & $\surd$ & $\surd$ &  & $\surd$ &  & 50.4 & 21.0 & - & - & - & - & - & - & - & - \\
L3MVN \cite{yu2023l3mvn}         & $\surd$ & $\surd$ & & $\surd$ & & 50.4 & 23.1 & 34.9 & 14.5 & - & - & - & - & - & - \\
VLFM  \cite{yokoyama2024vlfm}    & $\surd$ & $\surd$ & & $\surd$ & & 52.5 & 30.4 & 36.4 & 17.5 & 35.2 & 18.6 & 32.4 & 17.3 & 35.2 & 19.6 \\
GAMap \cite{huang2024gamap}      & $\surd$ & $\surd$ & & $\surd$ & & 53.1 & 26.0 & - & - & - & - & - & - & - & - \\
SG-Nav \cite{yin2024sg}     & $\surd$ & $\surd$ & & $\surd$ & & 54.0 & 24.9 & 40.2 & 16.0 & - & - & - & - & - & - \\
UniGoal \cite{yin2025unigoal}     & $\surd$ & $\surd$ & & $\surd$ & & 54.5 & 25.1 & 41.0 & 16.4 & - & - & - & - & - & - \\
CompassNav \cite{li2025compassnav} & $\surd$ & $\surd$ & & $\surd$ & & 56.6 & 27.6 & 42.0 & 17.5 & - & - & - & - & 43.5 & 21.6 \\
BeliefMapNav \cite{zhou2025beliefmapnav} & $\surd$ & $\surd$ & & $\surd$ & & 61.4 & 30.4 & 37.3 & \underline{17.6} & - & - & - & - & - & - \\
TriHelper \cite{zhang2024trihelper}      & $\surd$ & $\surd$ & & $\surd$ & & 62.0 & 25.3 & - & - & - & - & - & - & - & - \\
MTU3D \cite{zhu2025move}        & $\surd$ & $\surd$ & & $\surd$ & & - & - & - & - & 55.0 & 23.6 & 45.0 & 14.7 & 40.8 & 12.1 \\
WMNav \cite{nie2025wmnav}       & $\surd$ & $\surd$ & & $\surd$ & & 72.2 & 33.3 & 45.2 & 17.2 & - & - & - & - & - & - \\
CogNav \cite{cao2025cognav}     & $\surd$ & $\surd$ & & $\surd$ & & 72.5 & 26.2 & 46.6 & 16.1 & - & - & - & - & - & - \\

\midrule
NavFoM  \cite{zhang2025embodied}     & $\surd$ & & $\surd$ & & & - & - & - & - & 40.1 & 27.1 & 45.4 & \underline{32.6} & 45.2 & \underline{31.9} \\
Nav-$R^{2}$ \cite{xiang2025nav}       & $\surd$ &  & $\surd$ &  &  & - & - & - & - & 45.6 & 21.0 & 45.9 & 21.1 & 44.0 & 18.0 \\
Nav-R1  \cite{liu2025nav}    & $\surd$ & $\surd$ & $\surd$ & $\surd$ & $\surd$ & - & - & - & - & \underline{58.4} & 26.3 & 48.1 & 23.1 & 42.2 & 20.1 \\
zson    \cite{majumdar2022zson} & $\surd$ & & $\surd$ & & & 25.5 & 12.6 & 15.3 & 4.8 & - & - & - & - & - & - \\
Navid \cite{zhang2024navid}     & $\surd$ & & $\surd$ & & & 32.5 & 21.6 & - & - & - & - & - & - & - & - \\
PixNav \cite{cai2024bridging}    & $\surd$ & & $\surd$ & $\surd$ & $\surd$ & 37.9 & 20.5 & - & - & - & - & - & - & - & - \\
ESC \cite{zhou2023esc}           & $\surd$ & $\surd$ & $\surd$ & $\surd$ & & 39.2 & 22.3 & 28.7 & 11.2 & - & - & - & - & - & - \\
Uni-Navid  \cite{zhang2024uni}  & $\surd$ & & $\surd$ & & & \underline{73.7} & \underline{37.1} & - & - & 41.3 & 21.1 & 43.9 & 21.8 & 39.5 & 19.8 \\

\midrule
\textbf{\proj{}-Base (Stage 1)} & $\surd$ & & $\surd$ & $\surd$ & $\surd$ & 39.8 & 26.0 & 30.9 & 14.6 & 31.9 & 25.8 & 32.2 & 24.4 & 32.3 & 21.9 \\
\textbf{\proj{}-SFT (Stage 2)} & $\surd$ & & $\surd$ & $\surd$ & $\surd$ & 72.9 & 27.7 & \underline{49.0} & 16.5 & 57.2 & \underline{28.3} & \underline{56.9} & 27.2 & \underline{50.9} & 23.1 \\
\textbf{\proj{}-IRFT (Stage 3)} & $\surd$ & & $\surd$ & $\surd$ & $\surd$ & \textbf{84.8} & \textbf{41.1} & \textbf{64.0} & \textbf{29.6} & \textbf{65.0} & \textbf{34.4} & \textbf{63.9} & \textbf{35.1}  &  \textbf{66.3} & \textbf{37.4} \\
\bottomrule
\end{tabular}
}
\end{small}
\end{center}
\end{table*}

\begin{table*}[t]
\centering
\caption{\textbf{Model performance with different training recipes.} We analyze the effectiveness of different data strategies, progressive training pipeline, and memory configurations on HM3D and OVON Val-Unseen benchmark. The \textbf{best} and the \underline{second best} results are denoted by \textbf{bold} and \underline{underline}. Memory($L=k$) represents the memory with a maximum number of $k$ landmarks.}
\label{tab:unified_ablation}
\begin{small}
\resizebox{\textwidth}{!}{

\begin{tabular}{l c c c c c c | cc | cc}
\toprule
\multirow{2}{*}{\textbf{Ablation group}} & \multirow{2}{*}{\textbf{Exp \#}} & \multicolumn{5}{c}{\textbf{Module configuration}} & \multicolumn{2}{c}{\textbf{HM3D Val}} & \multicolumn{2}{c}{\textbf{OVON Val-Unseen}}\\

\cmidrule(lr){3-7} \cmidrule(lr){8-9} \cmidrule(lr){10-11}
 & & \textbf{Memory} & \textbf{Co-training} & \textbf{Stage 1} & \textbf{Expl.} & \textbf{S.P.} & \textbf{SR} & \textbf{SPL} & \textbf{SR} & \textbf{SPL} \\
\midrule

\multirow{3}{*}{Memory} 
 & 1 & w/o Memory & \checkmark & \checkmark & \checkmark & & 61.9 & 13.9 & 49.7 & 13.3 \\
 & 2 & Memory ($L=5$) & \checkmark & \checkmark & \checkmark & & \textbf{74.6} & \textbf{28.8} & \underline{50.0} & 22.3 \\
 & 3 & Memory ($L=15$) & \checkmark & \checkmark & \checkmark & & 70.8 & 28.1 & 47.5 & 22.4 \\

\midrule

\multirow{1}{*}{Co-training} 
 & 4 & Memory ($L=10$)  & & \checkmark & \checkmark & & 69.1 & 25.9 & 48.2 & 22.3 \\
\midrule

\multirow{1}{*}{Training Stage} 
 & 5 & Memory ($L=10$)  & \checkmark &  & \checkmark & & 68.0 & 24.6 & 41.8 & 14.1 \\
\midrule

\multirow{1}{*}{Data Collection Strategy} 
 & 6 & Memory ($L=10$)  & \checkmark & \checkmark & & \checkmark & 47.5 & \underline{28.7} & 40.4 & \textbf{28.1} \\
\midrule

\multirow{1}{*}{\textbf{\proj{} stage 2}} 
 & 7 & Memory ($L=10$)  & \checkmark & \checkmark & \checkmark & & \underline{72.9} & 27.7 & \textbf{50.9} & \underline{23.1} \\

\bottomrule
\end{tabular}
}
\end{small}
\end{table*}

\subsection{The overall performance of \proj{}}

\paragraph{The final performance of \proj{}-IRFT.}
\Cref{tab:sota_comparison} presents a comprehensive comparison between \proj{} and existing approaches. \proj{}-IRFT achieves new state-of-the-art (SOTA) results in SR and SPL across all settings, compared with methods that either output meta actions directly or rely on external tools for navigation.
On the HM3D validation set, \proj{}-IRFT achieves an SR of {84.8\%}, outperforming the previous SOTA Uni-Navid \cite{zhang2024uni} by {+11.1\%}. On the MP3D benchmark, which features larger scenes and more diverse object categories, \proj{}-IRFT improves over the previous SOTA CogNav \cite{cao2025cognav} by nearly {+17.4\%}.
The advantages of \proj{}-IRFT are most evident on the challenging OVON benchmark, designed to test the agent's generalization ability to synonyms and unseen object categories. On the Val-Synonyms split, \proj{}-IRFT shows a {+15.8\%} improvement over the previous SOTA Nav-R1 \cite{liu2025nav}. On the Val-Unseen split, which strictly evaluates zero-shot generalization, \proj{}-IRFT achieves an SR of {66.3\%}, significantly outperforming the previous SOTA NavFoM \cite{zhang2025embodied} (45.2\%) by {+21.1\%}.

\paragraph{Effectiveness of the curriculum training pipeline.} From \Cref{tab:sota_comparison}, we can observe that \proj{} achieves higher SR and SPL with more training stages. In particular, \proj{}-Base at stage 1 is inferior to previous SOTA, while \proj{}-SFT, after equipped \proj{} with reasoning and memory, shows significant improvement (33.1\% on the SR of HM3D) across all benchmarks. The performance of \proj{}-SFT is close to the previous SOTA on HM3D and achieves new SOTA on the MP3D and OVON-Val-Unseen. Furthermore, IRFT not only increases the search efficiency of \proj{} but also boosts the final success rate. Combining the efforts from all three stages, \proj{}-IRFT establishes highest SR and SPL across all settings, outperforming the second-best by a visible margin ({+21.1\%} on the SR of OVON-Val-Unseen).

\subsection{Ablation Studies}
\label{sec:ablation}

We perform an ablation study to analyze the contribution of different training strategies used in \proj{}. 
The results on HM3D Val and OVON Val-Unseen with different training setups are shown in \Cref{tab:unified_ablation}.

\paragraph{Effectiveness of memory.}
Comparing Exp 1, 2, 3, and 7, we observe that the model without memory (Exp 1) achieves reasonable SR but suffers from low SPL compared to the other three. This gap in SPL indicates that the memory module is crucial for search efficiency. Among Exp 2, 3, and 7, models with different number of landmarks in memory do not show significant differences in SR or SPL. We ultimately set the maximum number of landmarks to 10 for stage 2 due to its superior performance on the OVON Val-Unseen.

\paragraph{Impact of data collection strategy.}
In \Cref{sec:stage2}, we introduced a ``heuristic waypoint selection'' strategy to generate exploration trajectories (Expl.), rather than relying solely on the shortest path strategy (S.P.). Comparing Exp 6 with Exp 7, the model trained only on shortest paths (Exp 6) suffers significant degradation in SR, with a 25.4\% drop on HM3D and a 10.5\% drop on OVON Val-Unseen, compared to the model trained on trajectories with exploration (Exp 7). This indicates that exploration is a key capability for achieving high success rates. Interestingly, the model trained with shortest paths (Exp 6) achieves higher SPL compared to Exp 7. We attribute this to the fact that the model trained on shortest paths bypasses exploratory behavior, thereby achieving greater efficiency in straightforward scenarios where the target is in close proximity.

\paragraph{Importance of the progressive training pipeline.}
Our curriculum training pipeline consists of three distinct stages. Skipping stage 1 results in a noticeable performance drop in both SR and SPL (Exp 5 vs. Exp 7). Specifically, SR decreases by 4.9\%
 on HM3D and 9.1\% on OVON. Without the alignment between visual observations and meta-actions learned in stage 1, we observe the model sometimes fails to generate collision-free actions, causing it to get stuck more frequently, particularly in the unseen OVON environment.

\paragraph{Co-training improves performance} Comparing Exp 4 with Exp 7, removing co-training (Exp 4) leads to a slight drop in SR and SPL (e.g., 69.1\% vs. 72.9\% on HM3D). This suggests that co-training with general VQA data helps the VLM maintain its general image understanding capabilities, avoid overfitting to the specific navigation data distribution, and strengthen its ability to identify objects.

\begin{figure*}[t]  
    \centering      
    \includegraphics[width=0.9\textwidth,]{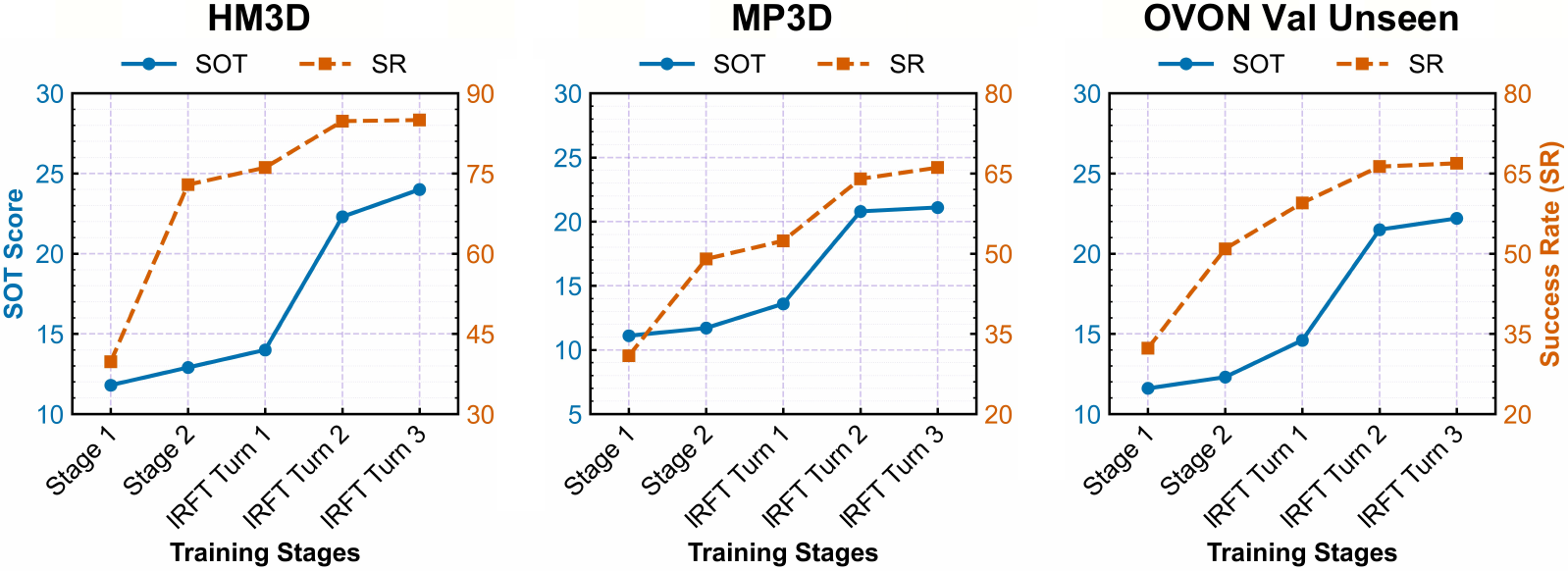}
    \caption{Performance analysis of multi-turn IRFT across different benchmarks.}
    \label{fig:multi-turn rft}
\end{figure*}

\subsection{Effectiveness of Multi-turn IRFT}

The evolution of SR and SOT across different training stages is presented in \Cref{fig:multi-turn rft}. We observe that \proj{} achieves higher SR and SOT as training progresses. In particular, \proj{}-Base (stage 1) shows limited performance, while stage 2 yields a significant improvement in SR across all benchmarks (e.g., $\uparrow 33.1\%$ on HM3D). This substantial gain confirms that the base VLM benefits from the temporal-spatial reasoning introduced in \Cref{sec:stage2}, which enables the agent to handle long-horizon exploration and memory. However, during stage 2, we notice that the improvement in SOT is marginal compared to that in SR. This is because \proj{}-SFT (stage 2) operates with a fixed reasoning frequency and therefore suffers from high inference costs in simple scenarios where reactive control suffices. Following stage 2, IRFT not only boosts the final success rate but also dramatically increases search efficiency in the first two turns. As we carry out more iterations of RFT, \proj{} achieves new state-of-the-art results across all settings. In particular, on OVON Val-Unseen, \projrft{} improves SOT from $12.3\%$ (stage 2) to $22.2\%$ (RFT turn 3). This demonstrates that the adaptive switching mechanism is effective: by learning from the ``reject-and-repair'' strategy, the agent learns to trigger the slow system only at critical ``stagnation points'' (e.g., lack of progress or repetitive exploration). This reduces unnecessary inference costs during the search process, resulting in a system that achieves both high SR and computational efficiency. We terminate IRFT at the third turn as we observe performance plateauing beyond this point. As detailed in \Cref{appendix:H}, compared to dense reasoning methods like VLMnav \cite{goetting2024end} and Nav-$R^{2}$ \cite{xiang2025nav}, \proj{}-IRFT achieves significantly higher SOT scores and SR by reducing the reasoning ratio to just $3.0\%$.

\subsection{Real-world Deployment}
\begin{figure*}[t]  
    \centering      
    \includegraphics[width=0.8\textwidth]{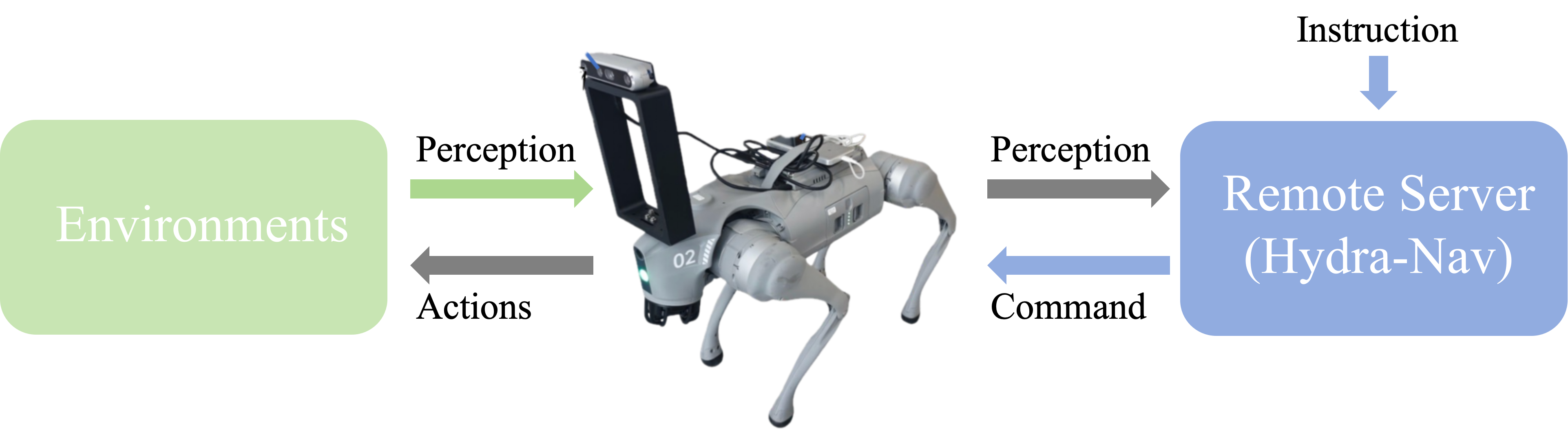}
    \caption{Real-world robot platform.}
    \label{fig:real robot}

\end{figure*}

\begin{figure*}[t]  
    \centering      
    \includegraphics[width=0.9\textwidth]{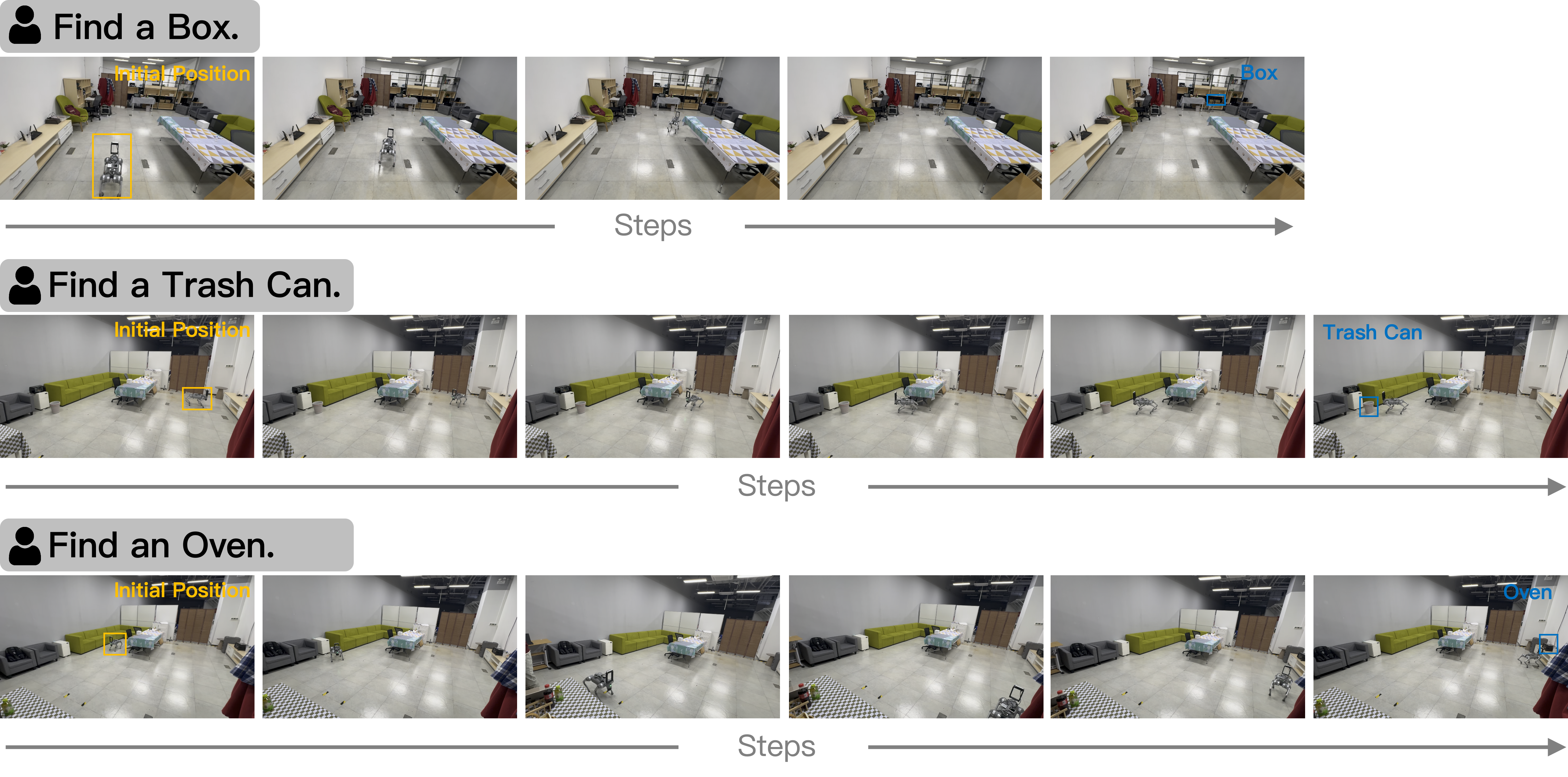}
    \caption{Qualitative performance of \proj{} in real-world deployments.}
    \label{fig:real robot experiment}
\end{figure*}

We deploy \proj{} directly on a physical robot without real-world fine-tuning. The model demonstrates robust zero-shot transfer, successfully locating target objects within complex environments. Our robot platform is shown in \Cref{fig:real robot}. The system is based on a Unitree Go2 quadruped, equipped with an Intel RealSense D457 camera (capturing $1280 \times 800$ RGB frames with a $90^{\circ}$ HFOV) and a portable Wi-Fi module for communication with a remote server. During deployment, the robot transmits compressed images ($640 \times 480$) to the server for next-action prediction. The robot executes the predicted action using a nonlinear model predictive control (NMPC) module \cite{grandia2023perceptive}, which computes optimal velocities based on a kinematic unicycle model. \Cref{fig:real robot experiment} illustrates real-world navigation episodes where the robot starts from an initial position and successfully locates a box, trash can, and oven.

\section{Limitations}
\label{sec:limitations}

A key limitation of this work is that we evaluate our model exclusively in the Habitat simulator. This is primarily due to the lack of high-quality evaluation sets in more realistic simulators such as IsaacSim. Future work could develop new benchmarks in more realistic simulations and investigate navigation models that generalize across different simulators and embodiments.

\section{Conclusion}

In this paper, we present \proj{}, a navigation agent that balances success rate with execution efficiency by unifying reactive control and deliberate reasoning within a single VLM. Our dual-process architecture enhances spatio-temporal reasoning and long-term memory, enabling the agent to switch adaptively between fast execution and deliberate reasoning. To optimize this trade-off, we employ iterative rejection fine-tuning (IRFT), teaching the agent to trigger reasoning only at critical stagnation points. Extensive experiments on standard benchmarks demonstrate that \proj{} achieves state-of-the-art performance with superior efficiency. 

\section{Future Works}

This work aims to advance the development of adaptive reasoning, vision-language model-based embodied agents for object navigation. While \proj{} achieves promising performance on existing object navigation benchmarks, several challenges remain. We outline three promising directions for future research:

\begin{itemize}

  \item Generalization to realistic environments: A key limitation of this work is that evaluations are conducted exclusively in the Habitat simulator due to the lack of high-quality benchmarks in more realistic environments like Isaac Sim. Future research should focus on developing high-fidelity benchmarks and investigating how well navigation models generalize across different simulators and, crucially, to the physical real world. 
  
  \item Optimizing adaptive switching mechanisms: While iterative rejection fine-tuning (IRFT) successfully teaches agents to reason at ``stagnation points'', future work could leverage online reinforcement learning. This would allow agents to autonomously learn the optimal timing for triggering reasoning and simultaneously improve the quality of the generated reasoning chains. 
  
  \item Expanding embodiment and task scope: Currently, \proj{} is tailored for object navigation. Future research could extend this dual-process architecture to a broader range of embodied tasks and robot types. This includes exploring how the ``slow-fast'' paradigm applies to more complex scenarios, such as mobile manipulation or autonomous game agents. 

\end{itemize}


\clearpage

\bibliographystyle{plainnat}
\bibliography{main}

\clearpage

\beginappendix

\section{The definition of the score function}
\label{sec:appendx_score_function}

Below is the score function used in \Cref{sec:stage2}
\[
    {S}(p_{\text{init}}, {p}) \coloneqq \underbrace{ \frac{ \min_{ q \in U } \| p - q \|_2 }{\max_{ q \in U } \| p - q \|_2} }_{ \text{spaciousness of $p$} } + \lambda \cdot \underbrace{\left( 1 - \frac{||{p} - {p}_{\text{target}}||_2}{\max_{q \in G} (|| {p}_{\text{init}} - q ||_2)} \right)}_{ \text{closeness to the target} },
\]
where $p$ represents the candidate point and $p_{\text{target}}$ is the location of goal. $U$ denotes the set of obstacle boundaries (e.g., walls). $G$ represents the set of all targets in the scene. The first term calculates the spatial openness by measuring the distance from $p$ to the nearest obstacle. The second term evaluates the proximity to the specific target, normalized by the distance from the start point $p_{\text{init}}$ to the farthest target in $G$. The weight $\lambda=0.7$ balances exploration against goal efficiency.

\section{The computation of operation time}
\label{detail for SOT}

In this section, we provide the specific parameter settings and calculation method used for the Success weighted by Operation Time (SOT) metric. The total operating time $T_{\text{actual}}$ consists of robot execution time and model inference latency, defined as follows:
$$T_{\text{actual}} = T_{\text{phys}} + T_{\text{inf}}$$

\paragraph{Physical Execution Estimation ($T_{\text{phys}}$).} We calculate robot execution time by summing meta action durations. To ensure realism, these durations represent the average values obtained from extensive real-world testing on the physical robot, as listed in Table 3.

\begin{table}[h]
\centering
\caption{Time cost parameters for robot physical actions.}
\vspace{2mm}
\label{tab:time_params}
\begin{tabular}{l|c|l}
\hline
\textbf{Action Type} & \textbf{Time Cost ($s$)} & \textbf{Description} \\ \hline
\texttt{MoveAhead} & 1.0 & Forward movement (approx. 0.25m) \\
\texttt{RotateLeft/Right} & 0.6 & In-place rotation (approx. 30$^\circ$) \\
\texttt{Obs} & 4.0 & Panoramic observation \\
\texttt{Stop} & 0.1 & Signal to terminate the episode \\ \hline
\end{tabular}
\end{table}

\paragraph{Inference Latency Estimation ($T_{\text{inf}}$).} The inference time is modeled as 
$$T_{\text{inf}} = \tau \times (N_{\text{CoT}} + N_{\text{action}}),$$ 
where $N_{\text{CoT}}$ and $N_{\text{action}}$ represent the token counts for reasoning and actions. We set $\tau = 0.015s$ based on real-time inference logs collected during navigation on a single NVIDIA H20 GPU.

\paragraph{Theoretical minimum physical time ($T_{\text{optimal}}$).} $T_{\text{optimal}}$ represents the physical execution time required for the robot to traverse the shortest path in the scene. 

\begin{figure*}[!htbp]  
    \centering   
    \includegraphics[width=0.5\textwidth]{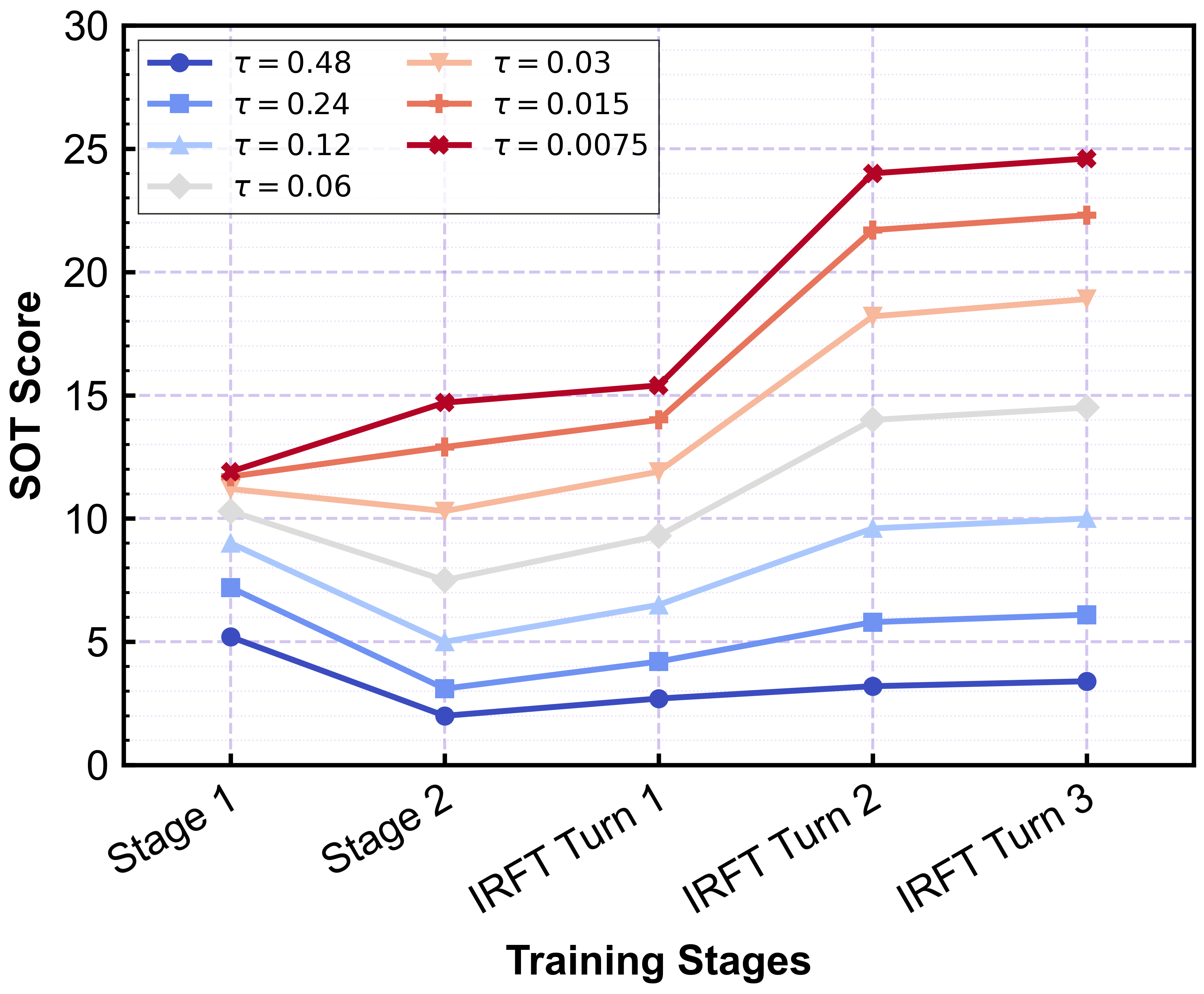}
    \caption{Performance visualization of \proj{}.}
    \label{fig:SOT_Multi_Tau}
\end{figure*}

\subsection{Sensitivity Analysis of SOT to Inference Latency \texorpdfstring{$\tau$}{tau}}
To address concerns regarding the dependency of SOT on hardware configurations, we conduct a rigorous sensitivity analysis of the SOT score with respect to the per-token inference latency $\tau$. We varied $\tau$ over a wide range, from $0.0075$s (simulating GPUs like H100) to $0.48$s (simulating high-latency edge scenarios). The results, summarized in \Cref{fig:SOT_Multi_Tau}, demonstrate the robustness of our adaptive reasoning model through three key observations:

\begin{itemize}

\item Consistent Superiority in Realistic Settings: Across the realistic range of inference latencies ($\tau \in [0.0075, 0.12]$), \proj{}-IRFT (Turn 3) consistently achieves the highest SOT score. This indicates that our efficiency gains are not the result of a specific $\tau$ value but are intrinsic to the improved policy.

\item {Resilience to High Latency via Adaptive Reasoning:} For models equipped with reasoning capability, IRFT proves highly resilient to hardware constraints. Comparing the adaptive reasoning model (IRFT) against the fixed-frequency reasoning model (stage 2), we observe that IRFT maintains a consistent upward trend in SOT across all latency conditions. Notably, as hardware becomes slower (increasing $\tau$), the performance gap widens in favor of IRFT. This confirms that by learning to trigger reasoning only at critical ``stagnation points'', IRFT maximizes the utility of compute, ensuring a net positive gain in efficiency even when inference costs are prohibitive.

\item {Boundary Analysis:} The stage 1 model (which lacks reasoning) only begins to show competitive SOT at extremely high latencies ($\tau \ge 0.24$), solely due to its minimal inference overhead. However, within the practical operating range of modern VLMs, the navigational benefits of reasoning vastly outweigh the time cost, provided that the reasoning is triggered adaptively as done in \proj{}.

\end{itemize}

\section{Comparison with SOTA baselines on SOT metric}
\label{appendix:H}

\subsection{Comparative SOT Results with SOTA Baselines}

To evaluate the trade-off between success rate and computational overhead, we compare \proj{} with three representative open-source, reproducible baselines: VLMnav \cite{goetting2024end}, Nav-$R^2$ \cite{xiang2025nav}, and WMNav \cite{nie2025wmnav}. As shown in \Cref{tab:sot_baseline}, we report the SR, SOT, and the reasoning ratio—defined as the percentage of steps where the reasoning is triggered. 

Existing VLM-based methods typically rely on dense reasoning, where the reasoning is triggered at every step, resulting in a $100\%$ reasoning ratio. Nav-$R^2$, an end-to-end method that directly outputs low-level actions, achieves a competitive SR of $65.0\%$ on HM3D. However, its SOT drops to just $1.9$, primarily because triggering reasoning for every meta action causes significant time delays. This highlights that high success rates are insufficient for practical deployment if search efficiency is low. VLMnav and WMNav adopt a hierarchical framework approach to mitigate this, achieving better SOT scores ($13.8$ and $17.9$ on HM3D). However, these hierarchical methods rely on the VLM acting as a brain to guide an ideal low-level executor. This dependency makes them difficult to deploy in the real world, where execution is imperfect. 

In contrast, \proj{}-IRFT demonstrates that frequent reasoning is often unnecessary. By learning to trigger the reasoning only at critical ``stagnation points'' via IRFT, our model maintains an extremely low reasoning Ratio of $3.0\%$ on HM3D and $3.5\%$ on OVON Val-Unseen. Despite this sparse reasoning, \proj{} outperforms the second-best method (WMNav) by substantial margins: $+12.1\%$ SR on HM3D and $+22.3\%$ SR on OVON Val-Unseen. Consequently, \proj{} achieves the highest SOT scores across all benchmarks. This empirically validates that unifying reactive low-level control with adaptive, low-frequency reasoning is key to balancing high success rates with search efficiency, while its unified nature facilitates direct deployment in the real world.

\begin{table*}[t]
\vspace{-2mm}
\centering
\caption{Comparison with open-source, reproducible methods on HM3D and OVON Val-unseen benchmarks. The attributes denote: \textbf{Low}: low-level action output, \textbf{High}: high-level planning/action output. The \textbf{best} and the \underline{second best} results are denoted by \textbf{bold} and \underline{underline}.}
\label{tab:sot_baseline}
\vskip 0.15in
\begin{small}
\setlength{\tabcolsep}{8pt} 
\begin{tabular}{l cc cc cc cc}
\toprule
\multirow{2}{*}{\textbf{Method}} & \multicolumn{2}{c}{\textbf{Action Level}} & \multicolumn{2}{c}{\textbf{HM3D Val}} & \multicolumn{2}{c}{\textbf{OVON Val-Unseen}} & \multicolumn{2}{c}{\textbf{\shortstack{Reasoning Ratio (\%)}}} \\
\cmidrule(lr){2-3} \cmidrule(lr){4-5} \cmidrule(lr){6-7} \cmidrule(lr){8-9}
& \textbf{High} & \textbf{Low} & \textbf{SR$\uparrow$} & \textbf{SOT$\uparrow$} & \textbf{SR$\uparrow$} & \textbf{SOT$\uparrow$} & \textbf{HM3D} & \textbf{OVON Val-Unseen} \\
\midrule

VLMnav \cite{goetting2024end} & \checkmark &  & 50.4 & 13.8 & 25.5 & 14.0 & 100.0 & 100.0\\
Nav-$R^{2}$ \cite{xiang2025nav} & & \checkmark & 65.0 & 1.9 & \underline{44.0} & 1.2 & 100.0 & 100.0\\
WMNav \cite{nie2025wmnav} & \checkmark & & \underline{72.2} & \underline{17.9} & 33.4 & \underline{16.9} & 100.0 & 100.0\\

\midrule




\textbf{\proj{}-IRFT Turn3} & & \checkmark & \textbf{84.3} & \textbf{24.0} & \textbf{66.3} & \textbf{20.3} & 3.0 & 3.5 \\

\bottomrule
\end{tabular}
\end{small}
\end{table*}

\begin{figure*}[htbp!]  
    \centering      
    \includegraphics[width=\textwidth]{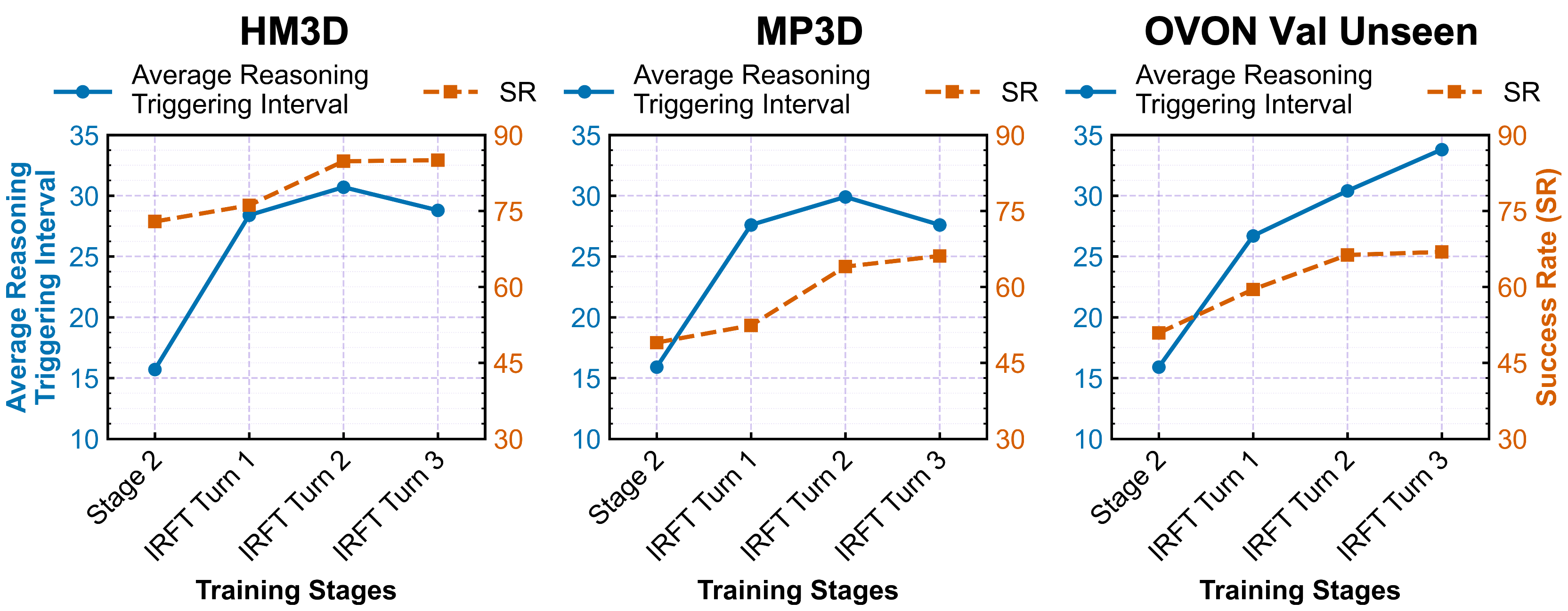}
    \caption{Reasoning Frequency analysis of multi-turn IRFT across different.}
    \label{fig:reasoning freq}
\end{figure*}

\subsection{Analysis of Reasoning Frequency}
\label{ap:Reasoning Frequency}
\Cref{fig:reasoning freq} shows the evolution of SR and reasoning frequency. Unlike stage 2, which relies on fixed, frequent reasoning (e.g., every $15.9$ steps on OVON Val-Unseen), IRFT reduces the reasoning frequency while boosting performance. By Turn 3, the reasoning interval on OVON Val-Unseen doubles alongside a significant increase in SR by $+16.0\%$. This confirms that the model learns to trigger the slow reasoning system only at critical ``stagnation points''. We attribute the slight fluctuations in reasoning frequency to the fact that as the agent improves and survives longer, it encounters harder scenarios that require occasional reasoning. We terminate IRFT at turn 3 as the SR stabilizes.

\section{Example of the long-term memory}
\label{Memory Example}
The long-term memory consists the following information:
\begin{itemize}
    \item \textbf{Landmark nodes:} Represented by the panoramic observations (e.g., ``At landmark1...''), where \texttt{<image>} tokens correspond to the 4 RGB images obtained by rotating the head camera at $90^\circ$ intervals ($0^\circ, 90^\circ, 180^\circ, 270^\circ$).
    \item \textbf{Action edges:} Represented by the execution meta actions connecting two consecutive landmarks.

\end{itemize}

\begin{tcolorbox}[
    colback=gray!10,      
    colframe=gray!50,     
    boxrule=0.5pt,        
    arc=2pt,              
    left=6pt, right=6pt, top=6pt, bottom=6pt,
    breakable,            
    title=\textbf{Memory Example} 
]

At landmark1, you see \texttt{<image>}\texttt{<image>}\texttt{<image>}\texttt{<image>}; 

Executed \texttt{RotateRight 30.0}, \texttt{MoveAhead 0.25}, \texttt{RotateRight 30.0}, \texttt{RotateRight 30.0}, \texttt{MoveAhead 0.25}, \texttt{RotateLeft 30.0}, \texttt{MoveAhead 0.25}, \texttt{MoveAhead 0.25}, \texttt{MoveAhead 0.25}, \texttt{RotateLeft 30.0}, \texttt{MoveAhead 0.25}, \texttt{RotateLeft 30.0}, \texttt{MoveAhead 0.25}, \texttt{MoveAhead 0.25}, \texttt{MoveAhead 0.25}, \texttt{MoveAhead 0.25}, \texttt{MoveAhead 0.25}, \texttt{MoveAhead 0.25}, \texttt{MoveAhead 0.25}, \texttt{MoveAhead 0.25}, \texttt{MoveAhead 0.25}, \texttt{MoveAhead 0.25}, \texttt{MoveAhead 0.25}, \texttt{RotateLeft 30.0}, \texttt{RotateLeft 30.0}, \texttt{RotateLeft 30.0}, \texttt{RotateLeft 30.0}, \texttt{RotateLeft 30.0} from landmark 1 to landmark 2; 

At landmark2, you see \texttt{<image>}\texttt{<image>}\texttt{<image>}\texttt{<image>}; 

Your current view is \texttt{<image>}.

\end{tcolorbox}

\section{Prompt for reasoning synthesis}
\label{Reasoning Prompt}
We adopt the following two steps to generate reasoning text.
\begin{itemize}
    \item Step 1: We prompt Qwen3-VL-235B-Thinking to summarize past visual observation along with the long-term memory.

    \item Step 2: Then we feed the summary from step 1, the current view, and the ``future correct view'' into the model. The model then generates reasoning text that serves to guide the future search plan.
\end{itemize}

The memory summary created in stage 1 is used as a key input for stage 2 to ensure the planning is based on history.

\begin{tcolorbox}[
    colback=gray!10,
    colframe=gray!50,
    boxrule=0.5pt,
    arc=2pt,
    left=6pt, right=6pt, top=6pt, bottom=6pt,
    breakable,
    title=\textbf{Step 1: Summarize History}
]
\small

\noindent {\#\#\# 1. Task Definition}
\begin{itemize}
    \setlength\itemsep{0em}
    \item You are an autonomous navigation agent.
    \item Your task is to analyze your visual history and raw memory log to consolidate your exploration history into a concise summary. You must verify which areas you have already explored.
\end{itemize}

\vspace{0.5em}
\noindent {\#\#\# 2. Inputs}
\begin{itemize}
    \setlength\itemsep{0em}
    \item {Visual Stream:} \texttt{[History\_Img\_1], ..., [History\_Img\_N]} \\
    (A full sequence of panoramic views from all previously visited landmarks).
    \item {Context:} You are provided with \texttt{\{number of images\}} images representing these views.
    \item {Raw Memory Log:} ``\texttt{\{action edges\}}''
\end{itemize}

\vspace{0.5em}
\noindent {\#\#\# 3. Output Format}
\begin{itemize}
    \setlength\itemsep{0em}
    \item Output {only} the summary paragraph.
    \item Do not include introductory phrases like ``Here is the summary''.
\end{itemize}

\end{tcolorbox}

\vspace{1.0em}

\begin{tcolorbox}[
    colback=gray!10,
    colframe=gray!50,
    boxrule=0.5pt,
    arc=2pt,
    left=6pt, right=6pt, top=6pt, bottom=6pt,
    breakable,
    title=\textbf{Step 2: Visual Hindsight Planning}
]
\small

\noindent {\#\#\# 1. Task Definition}
\begin{itemize}
    \setlength\itemsep{0em}
    \item {Role:} Navigation Agent.
    \item {Goal:} \texttt{\{goal\}}
    \item {Objective:} Identify actionable visual cues (e.g., doorways, corridors, objects) in the current view that logically lead to the scene shown in the future view. Formulate a high-level plan based on these cues and your memory.
\end{itemize}

\vspace{0.5em}
\noindent {\#\#\# 2. Inputs}
\begin{itemize}
    \setlength\itemsep{0em}
    \item {Current View (Images 1-4):} \texttt{[Img\_1], [Img\_2], [Img\_3], [Img\_4]} \\
    (A 360-degree panoramic observation of your current location).
    \item {Future View (Image 5):} \texttt{[Image\_5]} \\
    (The Ground Truth observation of the next step / Implicit Guide).
    \item {Memory Context:} ``\texttt{\{memory\_summary\}}'' \\
    (Consolidated history from Step 1).
\end{itemize}

\vspace{0.5em}
\noindent {\#\#\# 3. Output Format}
\begin{itemize}
    \setlength\itemsep{0em}
    \item Output the planning analysis and decision strictly in the following structure:
    \item {Visual Analysis:} ``Analyzing the current view, I see...''
    \item {High-level Plan:} ``Given that I have finished exploring [Area from Memory], and seeing [Feature] in the current view which likely leads to [Oracle Direction], I will proceed to...''
\end{itemize}

\vspace{0.5em}
\noindent {\#\#\# 4. Rules \& Constraints}
\begin{itemize}
    \setlength\itemsep{0em}
    \item {Hindsight Mechanism:} Image 5 reveals the correct future path.
    \item {Critical Constraint:} You must pretend you have {never seen} Image 5. You must rationalize your decision based {only} on the Current View (Images 1-4) and your memory, using Image 5 only as a hidden ground-truth guide.
    \item {Sequential Dependency:} You must explicitly reference the provided memory context to justify why you are choosing a new direction versus backtracking (e.g., confirming previous areas are fully explored).
\end{itemize}

\end{tcolorbox}

\section{System prompts}
\label{Agent Prompt}
We present the system prompts used for inference.

\begin{tcolorbox}[
    colback=orange!5,      
    colframe=orange!40,     
    boxrule=0.8pt,        
    arc=2pt,              
    left=6pt, right=6pt, top=6pt, bottom=6pt,
    breakable,            
    title=\textbf{System prompt for \proj{}-IRFT (stage 3)} 
]

\noindent {\#\# ROLE \& MISSION}

You are an expert autonomous navigation agent. You operate as a dual-process ``fast and slow'' thinking system to reach a user-specified goal.

\vspace{0.5em}
\noindent {``Slow System'' (Planning):} Activated at the start of the episode or following a self-initiated {obs} action. You will receive your full memory, the goal, and the {current\_view}. You must analyze this information to create a high-level plan.

\vspace{0.5em}
\noindent {``Fast System'' (Execution):} You execute your plan by generating low-level navigation actions based solely on your {current\_view}. You must remain in this mode until you reach the goal, determine a need for re-planning, OR reach the maximum execution limit (30 steps) since the last observation.

\vspace{0.8em}
\noindent {\#\# ACTION SPACE}

You must choose {only one} action from this list per turn. Do not output any text other than the single chosen action.

\begin{itemize}
    \setlength\itemsep{0em}
    \item {RotateLeft 30.0}: Rotate 30 degrees to the left.
    \item {RotateRight 30.0}: Rotate 30 degrees to the right.
    \item {MoveAhead 0.25}: Move forward 0.25 meters.
    \item {obs}: Stop and perform a 360-degree observation. This action reactivates the ``Slow System'' to update your memory and formulate a new plan.
    \item {end}: Terminate the mission when you are confident that you are within 1.0 meter of the goal.
\end{itemize}

\vspace{0.8em}
\noindent {\#\# CONVERSATION FLOW \& BEHAVIOR}

\vspace{0.5em}
\noindent {1. THE ``SLOW'' SYSTEM (Planning Mode)}
\begin{itemize}
    \setlength\itemsep{0em}
    \item {Input:} {Goal}, {Memory}, and {Panoramic View}.
    \item {Output Format:}
    
    \texttt{<think\_start>}
    
    Summarize memory...
    
    Analyze panoramic view...
    
    Formulate plan...
    
    \texttt{<think\_end>}
    
    [Action]
\end{itemize}

\vspace{0.5em}
\noindent {2. THE ``FAST'' SYSTEM (Execution Mode)}
\begin{itemize}
    \setlength\itemsep{0em}
    \item {Input:} only the {current\_view}.
    \item {Output Format:}
    
    [Action]
\end{itemize}

\vspace{0.8em}
\noindent {\#\# CRITICAL INSTRUCTIONS}

\begin{itemize}
    \setlength\itemsep{0.5em}
    \item {Task Completion:} You must output {end} exclusively when you are confident you are within 1.0 meter of the goal.
    
    \item {Adaptive Observation:} You should output {obs} if your path is blocked, the environment is ambiguous, or you are lost.
    
    \item {Forced Observation:} You must output {obs} if you have executed 30 consecutive actions in the ``Fast System'' without observing. The maximum interval between {obs} actions is strictly 35 steps.
\end{itemize}

\end{tcolorbox}

\section{Examples of trajectory}

\begin{figure*}[ht]  
    \centering      
    \includegraphics[width=\textwidth]{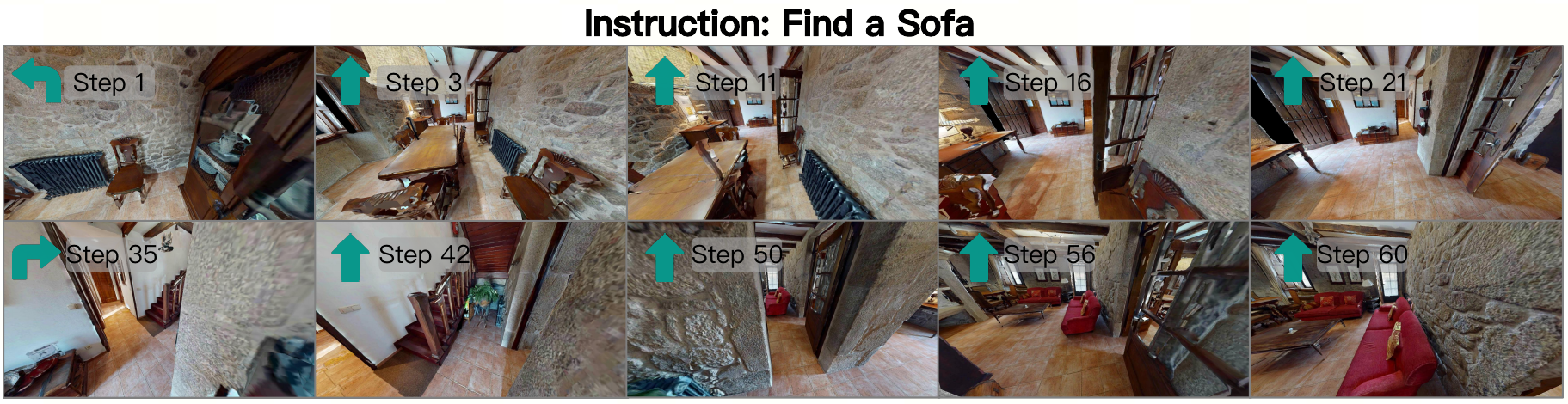}
    \caption{Performance visualization of \proj{}-Base.}
    \label{fig:stage1}

\end{figure*}

\begin{figure*}[ht]  
    \centering      
    \includegraphics[width=\textwidth]{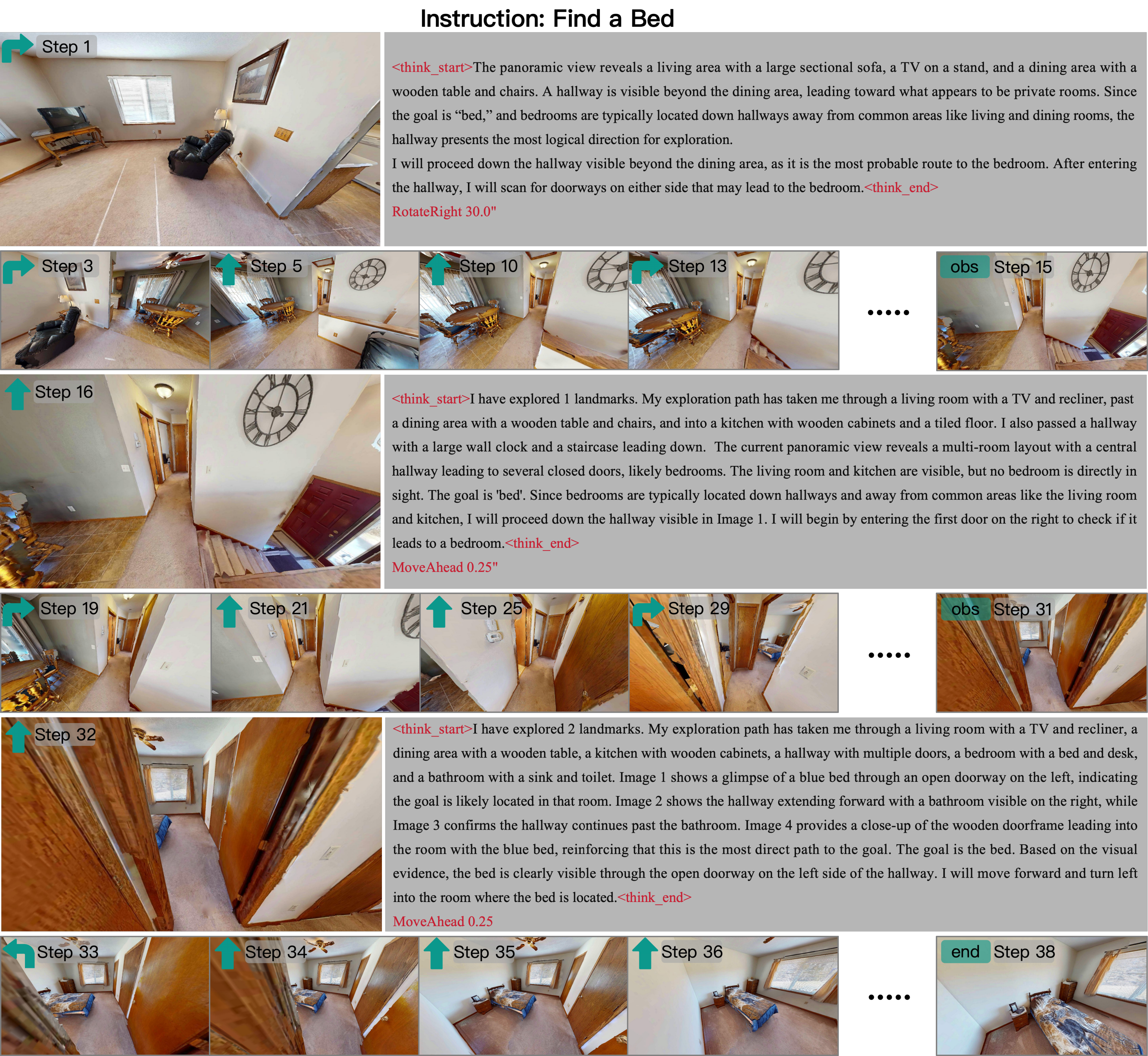}
    \caption{Performance visualization of \proj{}-SFT.}
    \label{fig:stage2}
\end{figure*}

\clearpage

\begin{figure*}[t!]  
    \centering      
    \includegraphics[width=\textwidth]{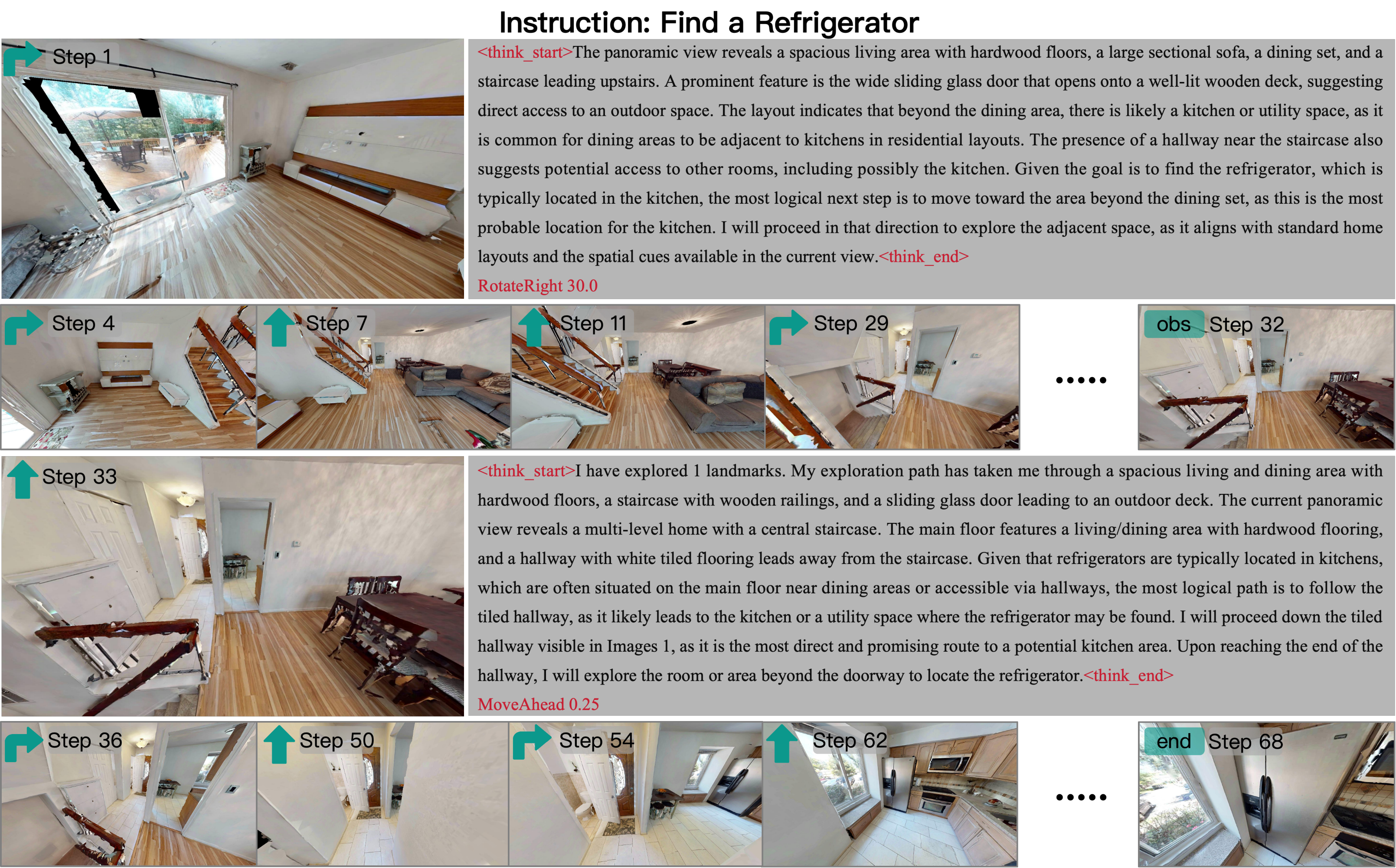}
    \caption{Performance visualization of \proj{}-IRFT.}
    \label{fig:stage3}
\end{figure*}

\section{Data collection for Stage 3}

\begin{minipage}[htbp]{0.48\textwidth}
\begin{algorithm}[H]
\caption{Rollouts with Stagnation Point Detection}
\label{alg:rollout_stage3}
\begin{algorithmic}[1]
\STATE \textbf{Input:} Goal object set $G$, Initial state $p_1$, Max steps $T_{\max} = 200, T_{\text{stag}}=20, \delta_{\text{stag}}=0.5m$
\STATE \textbf{Output:} Trajectory $\tau$

\STATE Initialize $\tau \leftarrow \emptyset$, $t \leftarrow 1$
\STATE $ForceSlow \leftarrow \textbf{True}$  \textit{// Start with slow system}

\WHILE{$t \leq T_{\max}$ \textbf{AND} (not $a_t = \texttt{End}$)}
    \STATE \textit{// 1. Check for stagnation point if in fast system}
    \IF{$ForceSlow = False$}
        \STATE {Condition 1 (Repetitive exploration):} Check if $\exists k \in [0, t-T_{\text{stag}}]$ s.t. $\| p_t - p_k \| \leq \delta_{\text{stag}}$
        \STATE {Condition 2 (Lack of progress):} Sample $\Delta t \sim \mathcal{U}(20, 35)$. Check if $\text{dist}(p_t, G) > \text{dist}(p_{t - \Delta t}, G)$

        \IF{Condition 1 is Met \textbf{OR} Condition 2 is Met}
            \STATE $ForceSlow \leftarrow \textbf{True}$ \textit{// Trigger reasoning (slow) system intervention}
        \ENDIF
    \ENDIF

    \STATE 
    \STATE \textit{// 2. Execute policy based on system mode}
    \IF{$ForceSlow = True$}
        \STATE \textit{// Slow system: reasoning + action}
        \STATE Input ``obs'' token to trigger reasoning
        \STATE Generate reasoning
        \STATE Sample action $a_t$ based on reasoning
        \STATE $ForceSlow \leftarrow \textbf{False}$ \textit{// Switch back to fast system immediately}
    \ELSE
        \STATE \textit{// Fast system: reactive meta action}
        \STATE Sample action $a_t$ directly from policy
    \ENDIF

    \STATE 
    \STATE \textit{// 3. Environment step}
    \STATE Execute $a_t$, observe $p_{t+1}$
    \STATE $\tau \leftarrow \tau \cup \{(p_t, a_t)\}$
    \STATE $t \leftarrow t + 1$
\ENDWHILE
\STATE \textbf{Return} $\tau$
\end{algorithmic}
\end{algorithm}
\end{minipage}
\hfill
\begin{minipage}[htbp]{0.48\textwidth} 
\begin{algorithm}[H]
\caption{IRFT Data Generation (Collection \& Repair)}
\label{alg:reapiar}
\begin{algorithmic}[2]
\STATE \textbf{Input:} Number of rollouts $N$, Goal object set $G$, Max steps $T_{\max}=200$, $\delta_{\text{success}}=1.0m$
\STATE \textbf{Output:} Training Dataset $\mathcal{D}$

\STATE \textit{// Phase 1: collect raw trajectories}
\STATE Initialize raw dataset $\mathcal{T}_{\text{raw}} \leftarrow \emptyset$
\FOR{$i = 1$ to $N$}
    \STATE Run \textbf{Algorithm 1} to collect trajectory $\tau$
    \STATE $\mathcal{T}_{\text{raw}} \leftarrow \mathcal{T}_{\text{raw}} \cup \{\tau\}$
\ENDFOR

\STATE 
\STATE \textit{// Phase 2: process and repair}
\STATE Initialize final training dataset $\mathcal{D} \leftarrow \emptyset$
\WHILE{$\mathcal{T}_{\text{raw}} \neq \emptyset$}
    \STATE Pop a trajectory $\tau$ from $\mathcal{T}_{\text{raw}}$
    
    \STATE \textit{// Check if the raw trajectory is successful}
    \IF{$a_T = \texttt{End} \land \text{dist}(p_T, G) \leq \delta_{\text{success}}$}
        \STATE $\mathcal{D} \leftarrow \mathcal{D} \cup \{\tau\}$
    \ELSE
        \STATE \textit{// Trajectory failed: identify intervention time $t^*$}
        \IF{$a_T = \texttt{End} \land \text{dist}(p_T, G) > \delta_{\text{success}}$}
            \STATE \textit{// Case 1: target misidentification}
            \STATE $t^* \leftarrow T$
        \ELSE
            \STATE \textit{// Case 2: timeout}
            \STATE Identify stagnation points $S \subset [1, T]$
            \STATE $t^* \leftarrow \operatorname*{argmin}_{t \in S} \text{dist}(p_t, G)$
        \ENDIF
        
        \STATE 
        \STATE \textit{// Perform repair strategy from $t^*$}
        \STATE Set action $a_{t^*} \leftarrow \texttt{obs}$ 
        \STATE Generate oracle path $\tau_{\text{oracle}}$ from $p_{t^*}$ to $G$ via $A^*$
        \STATE Synthesize reasoning for $\tau_{\text{oracle}}$
        \STATE $\tau_{\text{new}} \leftarrow \text{Concat}(\tau[1:t^*], \tau_{\text{oracle}})$
        
        \STATE \textit{// Check if repaired trajectory is successful}
        \IF{$|\tau_{\text{new}}| \leq 400$ \textbf{AND} $\text{dist}(p_T, G) \leq \delta_{\text{success}}$}
            \STATE $\mathcal{D} \leftarrow \mathcal{D} \cup \{\tau_{\text{new}}\}$
        \ENDIF
    \ENDIF
\ENDWHILE
\STATE \textbf{Return} $\mathcal{D}$
\end{algorithmic}
\end{algorithm}
\end{minipage}


\end{document}